\documentclass[lettersize,journal]{IEEEtran}
\usepackage{amsmath,amsfonts}
\usepackage{algorithmic}
\usepackage{algorithm}
\usepackage{array}
\usepackage[caption=false,font=footnotesize,labelfont=rm,textfont=rm]{subfig}
\usepackage{textcomp}
\usepackage{stfloats}
\usepackage{url}
\usepackage{verbatim}
\usepackage{graphicx}
\usepackage{cite}
\usepackage{multirow}%
\usepackage{bbding}
\usepackage{booktabs}
\newtheorem{remark}{Remark}
\begin{document}

\title{Camera Pose Refinement via 3D Gaussian Splatting} 

\author{Lulu Hao, Lipu Zhou,~\IEEEmembership{Member,~IEEE}, Zhenzhong Wei, Xu Wang
  }

\markboth{Journal of \LaTeX\ Class Files,~Vol.~14, No.~8, August~2021}%
{Shell \MakeLowercase{\textit{et al.}}: A Sample Article Using IEEEtran.cls for IEEE Journals}

\IEEEpubid{0000--0000/00\$00.00~\copyright~2021 IEEE}

\maketitle

\begin{abstract}
    Camera pose refinement aims at improving the accuracy of initial pose estimation for applications in 3D computer vision. Most refinement approaches rely on 2D-3D correspondences with specific descriptors or dedicated networks, requiring reconstructing the scene again for a different descriptor or fully retraining the network for each scene. Some recent methods instead infer pose from feature similarity, but their lack of geometry constraints results in less accuracy. To overcome these limitations, we propose a novel camera pose refinement framework leveraging 3D Gaussian Splatting (3DGS), referred to as GS-SMC. Given the widespread usage of 3DGS, our method can employ an existing 3DGS model to render novel views, providing a lightweight solution that can be directly applied to diverse scenes without additional training or fine-tuning. Specifically, we introduce an iterative optimization approach, which refines the camera pose using epipolar geometric constraints among the query and multiple rendered images. Our method allows flexibly choosing feature extractors and matchers to establish these constraints. Extensive empirical evaluations on the 7-Scenes and the Cambridge Landmarks datasets demonstrate that our method outperforms state-of-the-art camera pose refinement approaches, achieving 53.3\% and 56.9\% reductions in median translation and rotation errors on 7-Scenes, and 40.7\% and 53.2\% on Cambridge.
\end{abstract}

\begin{IEEEkeywords}
    3D Gaussian Splatting, camera pose refinement, multi-view constraints, visual localization.
\end{IEEEkeywords}

\section{Introduction}
\IEEEPARstart{V}{isual} localization, which estimates the six-degree-of-freedom (6-DoF) camera pose of a query image within a known scene, has a wide range of applications, including Augmented/Virtual Reality (AR/VR) \cite{WebAR_2025}, \cite{tvcg_ar_2022}, Simultaneous Localization and Mapping (SLAM) \cite{photometric_2022}, \cite{tvcg_slam_2024}, and autonomous navigation \cite{TPAMI_navigation_semantic_2025}, \cite{TRO_navigation_feature-matching_3dgs_2025}. Over the past decade, traditional structure-based localization has represented scenes as sparse 3D point clouds generated via Structure-from-Motion (SfM), retrieving 2D-3D correspondences from either handcrafted feature descriptors \cite{colmap_2016}, \cite{hybird_sfm} or learned feature descriptors \cite{hloc_2019}, \cite{TIP_sematics_feature_loc_2024}. Camera pose is then estimated using the RANSAC-P$n$P algorithm \cite{perspective_pnp}, \cite{ransac}. However, these approaches are tied to the specific descriptors used for scene reconstruction, which are inflexible and incur significant storage overhead. 

While structure-based methods rely on point clouds, regression-based methods use neural networks represent scenes implicitly. Scene Coordinate Regression (SCR) methods \cite{DSAC_2021}, \cite{ACE_2023}, \cite{GLACE_2024} predict the 3D coordinates of each image pixel, which are computationally expensive and impractical for large-scale or highly complex scenes. On the other hand, Absolute Pose Regression (APR) methods \cite{atloc_2020}, \cite{ms-transformer_2021}, \cite{dfnet}, \cite{lens} directly infer 6-DoF camera pose from the query image through end-to-end networks. Although APR methods achieve fast inference, their accuracy is highly dependent on the training data. Moreover, their performance typically degrades significantly under variations in illumination, viewpoint, or scene appearance.

Considering the above challenges, camera pose refinement provides a complementary approach to visual localization \cite{survey_relocalization_2024}, aiming to improve the accuracy of initial poses estimated by approaches such as APR. While some methods \cite{pixloc_2021}, \cite{poco_2021} refine the camera pose by aligning deep features between the query and reference images, they still depend on 2D-3D correspondences obtained from SfM point clouds. Recently, Neural Radiance Fields (NeRF) \cite{nerf_orgin} have attracted significant attention for their ability to provide an implicit representation of 3D scenes. Early NeRF-based refinement methods \cite{inverting_inerf_2021}, \cite{inverting_parallel_nerf_2023} invert a NeRF model to iteratively optimize the camera pose by aligning pixel-level intensities between the query and the rendered image (i.e., rendered by the NeRF model from the initial poses) using photometric error. Subsequent studies \cite{crossfire}, \cite{NeFeS50}, \cite{cs-hr-apr} employ tailored deep features to align the query image with the rendered image, but remain constrained by the need to sample a single ray per pixel during NeRF rendering, resulting in high computational cost. With the emergence of 3D Gaussian Splatting (3DGS) \cite{3dgs_orgin}, \cite{tvcg_3dgs_review_2024}, \cite{tvcg_3dgs_sky_2025}, which offers an explicit scene representation, researchers have increasingly leveraged 3DGS for camera pose refinement. Current 3DGS-based refiners either rely on dedicated neural networks that learn scene-specific features for pose refinement \cite{inverting_gsloc_2024}, \cite{TMECH_3DGS_2024}, or follow structure-based pipelines that establish 2D-3D correspondences using specific feature descriptors \cite{gscpr}.

\begin{figure}[!t]
    \centering
    \subfloat[\label{fig:other_refinement}]{\includegraphics[width=1\linewidth, trim=1 2 1 1, clip]{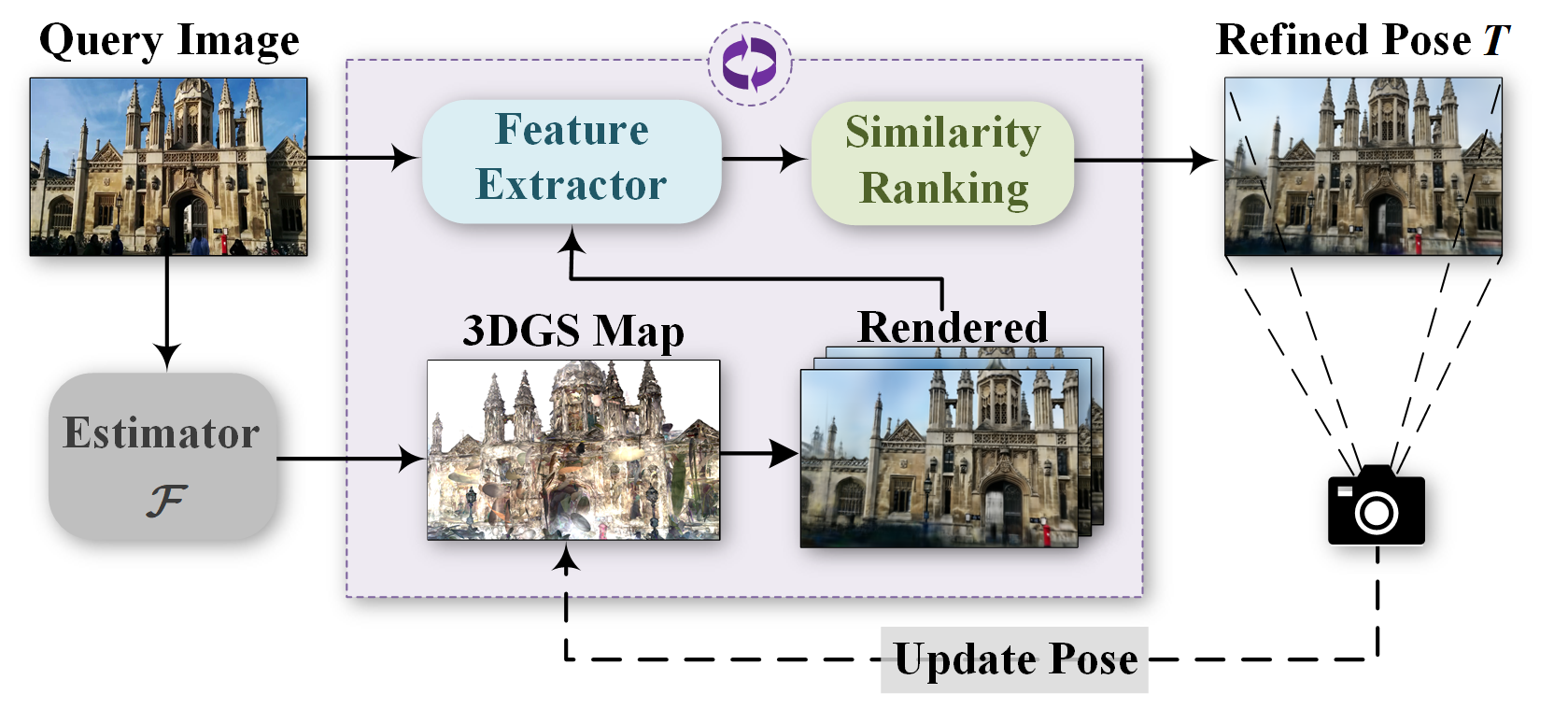}}\\
    \vspace{-0.3cm}
    \subfloat[]{\includegraphics[width=1\linewidth, trim=10 10 30 20, clip]{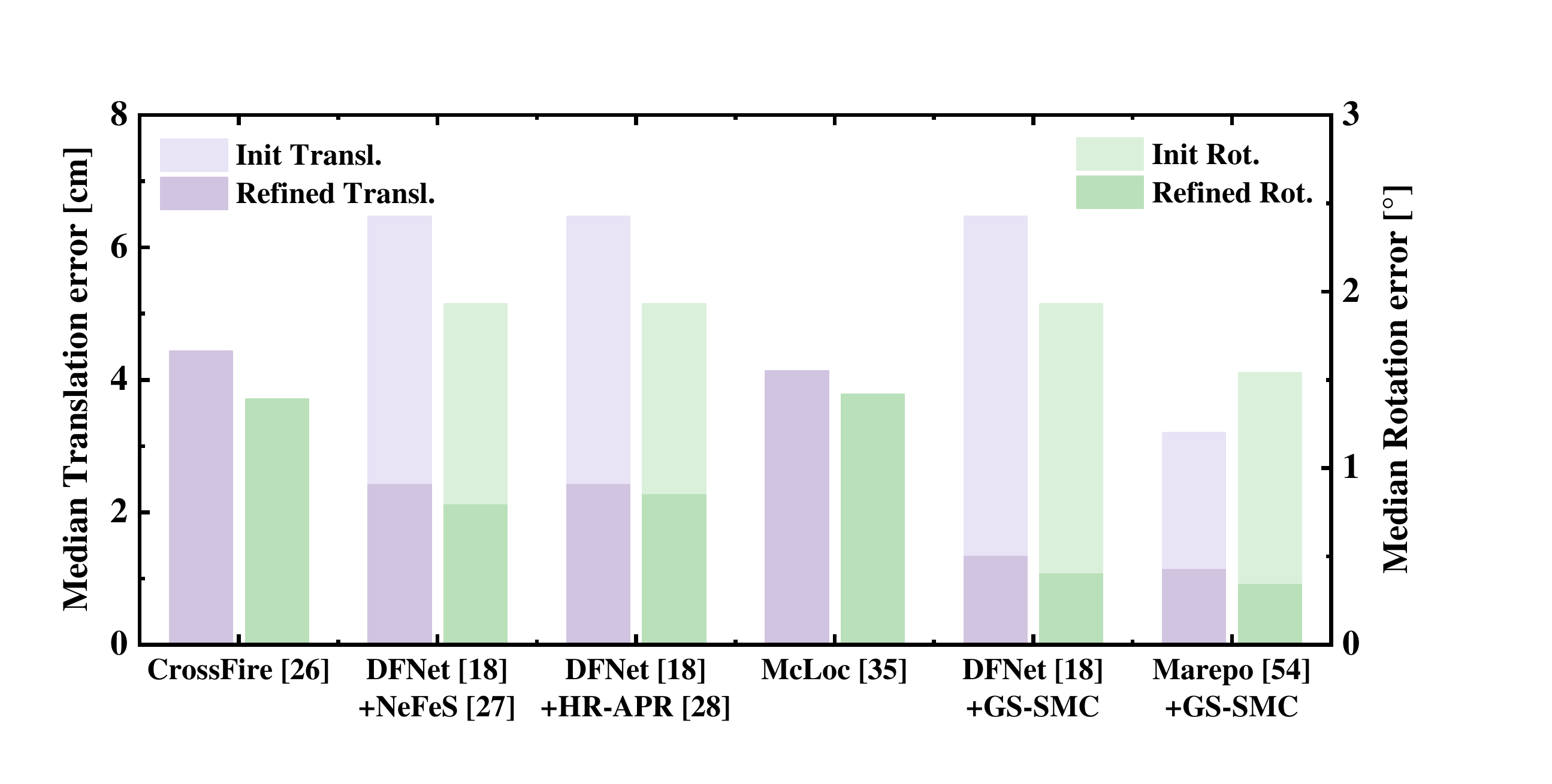}}
    \caption{(a) Existing pose refinement methods that do not require training on dedicated neural networks \cite{mcloc}, \cite{Transactions_uav_refinement_3dgs_2025}, generally adopt an iterative framework that utilizes feature-level similarity metrics to rank candidate poses. (b) Average median translation and rotation errors on the 7-Scenes dataset, where our method (GS-SMC) outperforms state-of-the-art approaches.}
    \label{fig:compare}
\end{figure}

\IEEEpubidadjcol 
Despite their promising performance, these approaches are limited by their reliance on either task-specific networks or descriptor types. Recent approaches \cite{mcloc}, \cite{Transactions_uav_refinement_3dgs_2025} generate multiple rendered images from perturbed poses and compute feature-level similarity between the query and rendered images to rank candidate poses, as shown in Fig.~\ref{fig:compare}\subref{fig:other_refinement}. However, they often neglect geometric information critical for precise pose refinement, limiting the potential of pre-trained features in the pose refinement task. While a straightforward method involves refining the camera pose using depth information from the 3DGS map combined with feature matching and the RANSAC-P$n$P algorithm, it exhibits shortcomings due to reconstruction quality.

To this end, we propose a novel camera pose refinement framework that iteratively refines the camera pose, named GS-SMC. Our approach is applicable across diverse scenes and is not tied to specific feature descriptors or dedicated neural networks. We employ 3D Gaussian Splatting (3DGS) for explicit scene representation and high-fidelity novel view synthesis. The image rendered from the initial pose is defined as the reference image, and the images rendered from perturbed poses generated by applying perturbations to the initial pose are defined as candidate images. Given these images along with the query image, our approach allows flexible choice of feature extractors and matchers to establish 2D-2D correspondences. Based on these matches, we iteratively refine the camera pose by computing the relative poses between the query and rendered images, leveraging epipolar geometry constraints. Additionally, our method can be seamlessly integrated into existing pose estimation frameworks to improve their localization performance further.

The main contributions of this paper are summarized as follows:
\begin{itemize}
    \item[$\bullet$] We propose a novel camera pose refinement framework that iteratively optimizes camera pose by matching the query image with novel views rendered from 3DGS. As 3DGS is widely adopted in 3D reconstruction and novel view synthesis, we can directly utilize an existing 3DGS model, providing a lightweight and adaptable solution for diverse scenes.
    \item[$\bullet$] We introduce an iterative method that refines the initial camera pose by estimating relative poses computed from 2D-2D correspondences between the query and rendered images via epipolar geometry. Our approach supports arbitrary feature extractors and matchers to establish these correspondences, avoiding time-consuming repeat reconstruction or full retraining per-scene networks.
    \item[$\bullet$] We conduct extensive evaluations on the 7-Scenes and Cambridge Landmarks datasets. The results indicate that our method significantly outperforms state-of-the-art refinement approaches, reducing the median translation and rotation errors by 1.29 cm and 0.45$^\circ$ on the 7-Scenes, and by 0.11 m and 0.33$^\circ$ on the Cambridge.
\end{itemize}

\section{Related Work}

In this section, we first present a brief review of existing approaches related to visual localization, then discuss recent advances in NeRF-based and 3DGS-based methods, and finally focus on the camera pose refinement and image alignment, which are most relevant to the proposed method. 

\subsection{Visual Localization}
Traditional structure-based methods \cite{hybird_sfm}, \cite{hloc_2019}, \cite{AS}, \cite{tvcg_bow_2023} rely on 2D-3D correspondences coupled with specific feature descriptors (i.e., SIFT \cite{sift}). Subsequent works have developed these structure-based techniques within deep-learning frameworks, making the SfM pipeline robust \cite{deeplearning_sfm_revisited_2021}, long-term \cite{TIP_sematics_feature_loc_2024}, and efficient \cite{deeplearning_sfm_vggsfm_2024}. Despite their superior performance, these methods depend on a sparse point cloud and a specific feature descriptor type, requiring repeated reconstruction for a new scene or when changing descriptor type.

Furthermore, regression-based methods \cite{review_deep_learing_visual_localization_2023} have emerged as an alternative to traditional methods that depend on sparse point clouds. Scene Coordinate Regression (SCR) predicts dense 3D coordinates for each image pixel and estimates the camera pose using differentiable optimization \cite{pixloc_2021}, \cite{uncertainty_scr_2016}, \cite{scr_Learning_less_is_more_2018}. Specifically, DSAC \cite{DSAC_2021} comprised two separate CNNs to predict scene coordinates, followed by a differentiable RANSAC algorithm for probabilistic hypothesis selection. ACE \cite{ACE_2023} utilized a scene-independent feature backbone and a scene-specific prediction head, yet still necessitates fine-tuning for mapping. GLACE \cite{GLACE_2024} integrated pre-trained global and local encodings, implicitly grouping the reprojected constraints according to co-visibility. These methods demand large amounts of training data and face challenges in large-scale scenes. Alternatively, Absolute Pose Regression (APR) predicts the camera pose directly from the query image using an end-to-end neural network \cite{posenet_cambridge_2015}, \cite{geometric_posenet_2017}, \cite{pose_lstm_2017}. ATLoc \cite{atloc_2020} and MS-Transformer \cite{ms-transformer_2021} improved APR by extracting salient feature points from feature maps and incorporating self-attention mechanisms based on the Transformer architecture. However, their accuracy largely also depends on the quantity and quality of training data, and they often degrade under varying illumination and viewpoints.

\begin{figure*}[!t]
    \centering
    \includegraphics[page=1, width=1\linewidth, trim=0 1 0 0, clip]{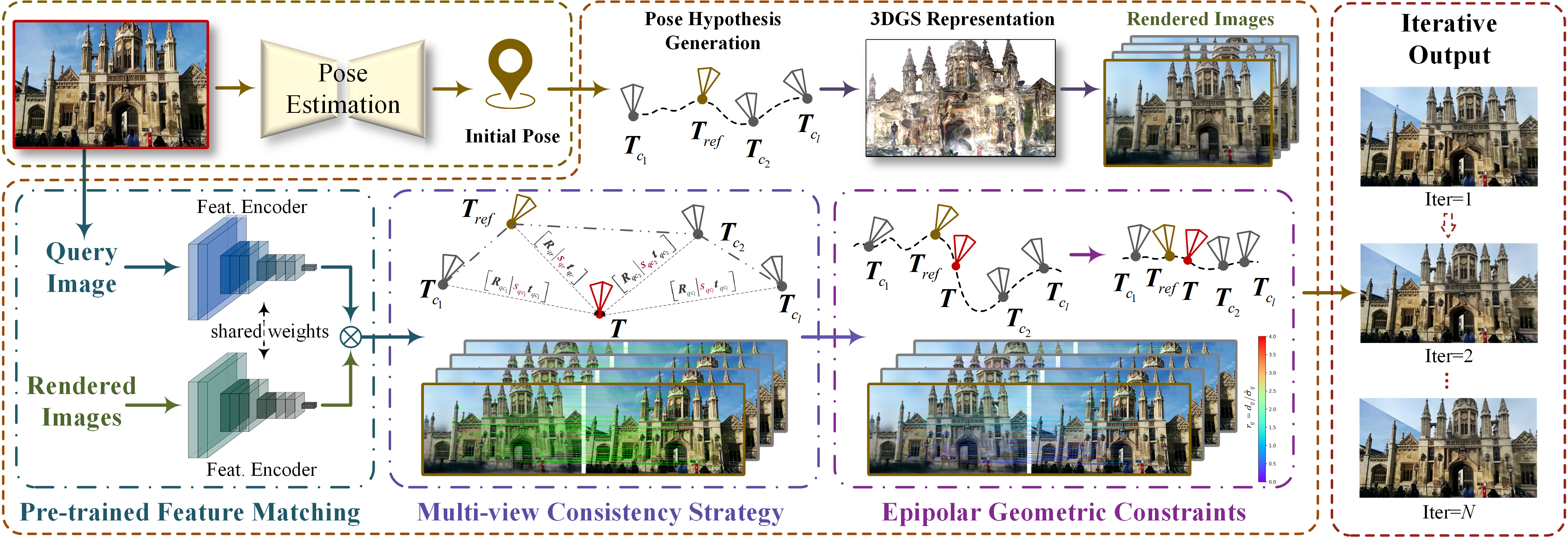}
    \caption{Overview of the proposed GS-SMC camera pose refinement method. Given a query image and its initial pose from prior estimation, we generate a set of pose hypotheses by applying perturbations to the initial pose and employ 3DGS to render images. We introduce a consistency strategy to refine the pose by estimating the relative poses between the query image and the rendered images. To further improve robustness, epipolar geometry constraints are introduced by minimizing the objective function \(\mathcal{L}\) at each iteration. The refinement proceeds iteratively, with each output displaying the ground truth in the bottom-left and the current prediction in the top-right.}
    \label{fig:overview}
\end{figure*}

\subsection{NeRF/3DGS-based Visual Localization} 

In recent years, novel view synthesis techniques have been developed to improve localization performance. Neural Radiance Fields (NeRF) represent scenes implicitly by modeling the radiance field with neural networks \cite{nerf_orgin}, \cite{tvcg_nerf_Vox_2022}, \cite{tvcg_nerf_motionaware_2024}. Previous studies estimate pose by inverting pre-trained NeRF models via pixel sampling \cite{inverting_inerf_2021}, \cite{inverting_nerf_APR_2021}, \cite{nvcg_itooth_2023}. Loc-NeRF \cite{inverting_mcnerf_2023} and Lin et al. \cite{inverting_parallel_nerf_2023} employed Monte Carlo sampling to alleviate convergence to local optima. Other methods advance Absolute Pose Regression (APR) by incorporating synthetic views generated from NeRF as additional training data \cite{dfnet}, \cite{lens}. Subsequent approaches focus on modeling viewpoint-dependent descriptors \cite{FQN}, dense self-supervised local features enriched with depth information \cite{crossfire}, \cite{cs-hr-apr}, and encoding 3D geometric features \cite{NeFeS50}. However, these approaches demonstrate limited scalability in large-scale scenarios. Recently, Marepo \cite{cs-marepo} introduced a scene-agnostic pose regression method adaptable to new map representations, though it still requires fine-tuning.

Although NeRF-based approaches advance visual localization, they are challenged by the requirement to sample a single ray per pixel. In contrast, 3D Gaussian Splatting (3DGS), which explicitly represents scenes using numerous 3D Gaussian ellipsoids, has attracted significant attention  \cite{3dgs_orgin}, \cite{sca-gs}, \cite{tvcg_3dgs_review_2024}. Early studies focus on inverting 3DGS models to achieve accurate camera pose estimation \cite{inverting_3dgs_icomma_2023}, \cite{inverting_gsloc_2024}, \cite{inverting_hgsloc_2024}, \cite{Neural_IPS_motiongs}. Zhou et al. \cite{TMECH_3DGS_2024} exploited photometric constraints and feature correspondences for precise pose estimation. While these early 3DGS-based approaches achieve notable pose estimation accuracy, they often rely heavily on photometric consistency and are sensitive to texture variations. More recently, NuRF \cite{Trans_particle_filter_3dgs_2025} proposed an adaptive nudged particle filter framework, though it still faces challenges in texture-poor environments. SplatLoc \cite{tvcg_splatloc_2025} introduced an unbiased 3D descriptor field for Gaussian primitives, but still faces difficulties when scaling to large outdoor scenes. Besides, SplatPose \cite{splatpose_2025} combined a dual-branch network with dense 2D-3D feature alignment, addressing feature misalignment and depth errors.

While these methods achieve significant improvements, our approach provides complementary and practical benefits for visual localization by refining the initially estimated camera poses. Additionally, by leveraging an existing pre-trained 3DGS model, it enables flexible adaptation to new scenes without the need for fully retraining or fine-tuning.

\subsection{Pose Refinement and Image Alignment} 

Pose refinement improves the accuracy of the initial pose estimate through the iterative minimization of a carefully designed objective function. Inspired by techniques widely employed in Simultaneous Localization and Mapping (SLAM) systems \cite{photometric_2022}, \cite{DSO_directi_align_2017}, \cite{tvcg_visualslam_2017}, direct alignment methods \cite{pixloc_2021}, \cite{Lm-reloc_2020}, \cite{Gn-net_2020} typically optimize pose estimation by aligning deep features with image gradients extracted from the query image and the current pose estimate, relying on the photometric constancy assumption. In contrast, indirect alignment methods, which rely on explicit feature matching, instead aim to minimize the reprojection error \cite{geometric_posenet_2017}, \cite{indirect_2024}, \cite{LoD-Loc_2025}, offering greater robustness.

Lately, NeRF/3DGS-based pose refinement methods have demonstrated notable improvements. \cite{inverting_parallel_nerf_2023}, \cite{inverting_inerf_2021}, \cite{inverting_nerf_APR_2021}, \cite{Neural_IPS_motiongs} performed gradient-based optimization by backpropagating the photometric error and refining camera pose using the Adam optimizer \cite{adam}. Alternatively, indirect alignment methods either design tailored features \cite{NeFeS50}, \cite{FQN}, \cite{crossfire}, \cite{cs-hr-apr}, \cite{Self-Supervised_nerf_2024_3dv} or rely on specific descriptors \cite{trans_atloc_2024}, \cite{gscpr} to encode the mapping between image features and camera pose. However, while the former requires task-specific training, the latter depends on the reconstruction pipeline, leading to limited practical applicability. Recent research investigates whether task-specific training for each scene is essential. One such method is Geo-Loc \cite{Transactions_uav_refinement_3dgs_2025}, which ranks retrieved database images using off-the-shelf features combined with a view-consistency-guided fusion module. Likewise, MCLoc \cite{mcloc} integrated pre-trained features with a particle filter framework to refine pose estimates iteratively.

Despite recent methods eliminate the need for training specific feature extractors and scene representation by ranking feature-level similarity, they remain limited in capturing geometric information useful for visual localization. In contrast, our method formulates camera pose refinement as a geometry-guided multi-view constrained optimization, allowing flexible selection of feature extractors and matchers.

\section{Methodology} 

We propose a camera pose refinement framework, termed GS-SMC, which leverages 3DGS to render novel views. In this section, we first define the camera pose refinement problem (Sec. \ref{3.1}). Next, we present the multi-view consistency strategy, which computes the refined pose by estimating the relative poses between the query image and the rendered images (Sec. \ref{3.2}). Finally, we introduce an epipolar geometry-constrained pose optimization approach, which further reduces the impact of mismatches in each refinement iteration (Sec. \ref{3.3}).

\begin{figure*}[!t]
    \centering
    \includegraphics[page=1, width=1\linewidth, trim=1 20 0 0, clip]{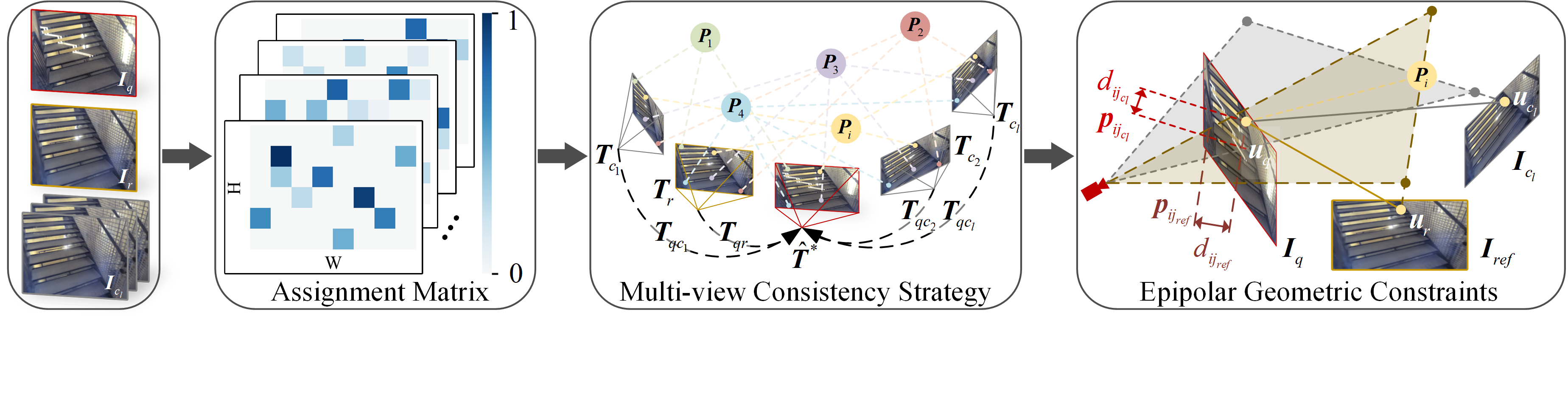}
    \caption{Illustration of multi-view constraints. In the first step, feature correspondences between the query image and rendered images are established based on the assignment matrix. The second step introduces a multi-view consistency strategy by estimating the relative poses and their corresponding motion relationships. In the third step, the camera pose is refined by minimizing the sum of distances from feature points to their corresponding epipolar lines. }
    \label{fig:illustrated_part}
\end{figure*}

\subsection{Problem Formulation}\label{3.1}

Since most visual localization and 3D reconstruction methods utilize 3DGS as the scene representation, our method can directly leverage an existing 3DGS model for pose refinement without additional full retraining or fine-tuning. In 3DGS, the scene is characterized by a dense set of anisotropic 3D Gaussians ${\cal G}\left(\boldsymbol{\mu}, \boldsymbol{\mit{\Sigma}}, {c}\left( \boldsymbol{d} \right) ,\alpha\right)$, where center position \(\boldsymbol{\mu}\in {{\mathbb{R}}^{3}}\), symmetric covariance matrix \( \boldsymbol{\mit{\Sigma}} \in {{\mathbb{R}}^{3\times 3}}\), spherical harmonics ${{{c}} \left( \boldsymbol{d} \right)}$ related to view direction \(\boldsymbol{d}\), and opacity ${{\alpha}}\in {{\mathbb{R}}^{1}}$. Let \(\boldsymbol{o}\in {{\mathbb{R}}^{3}}\) represent the center of the camera optical, the RGB color of each pixel \(\boldsymbol{I}(\boldsymbol{o}, \boldsymbol{d})\) on the image plane is computed by the alpha-blending formula.

Therefore, this paper focuses on refining the camera pose with a query image \({{\boldsymbol{I}}_{q}}\), a known camera pose \(\boldsymbol{T}\in \operatorname{SE}\left( 3 \right)\), and a trained \(3\)D representation of the scene parameterized by \(\cal G\) with known camera intrinsic matrix \(\boldsymbol{K}\). A straightforward approach could leverage 3DGS to render images and aligned depth maps, followed by establishing 2D-3D correspondences through feature matching for pose refinement using the RANSAC-P$n$P algorithm. However, the accuracy of this method is highly dependent on the quality of the 3DGS model and the robustness of feature matching, which significantly impacts the reliability of the derived depth information.

Consequently, we formulate pose refinement as an optimization problem that computes the pose of the query image from estimated the relative poses between the query image and the rendered images. An overview of the proposed method is presented in Fig.~\ref{fig:overview}, which follows the iterative approach. The refinement process is formulated as follows:
\begin{equation}
{\boldsymbol{\hat T}^*} = \mathop {{\mathop{\text{arg min}}\nolimits} }\limits_{{\boldsymbol{T}} \in {\mathop{\rm SE}\nolimits} \left( 3 \right)} {\mkern 1mu} {\cal L}({\boldsymbol{T}}\left| {{\boldsymbol{I}}_q}, {\cal G}, \boldsymbol{K},\delta _R^l, \delta _t^l\right.),
\end{equation}
where $\delta_R^l$ and $\delta_t^l$ represent the rotation and translation perturbations applied to $\boldsymbol{T}$, generating $L$ hypothetical poses, indexed by $l = 1, \dots, L$.

In addition, our implementation can be expressed as follows:
\begin{equation}
{\boldsymbol{\hat T}^*} = \mathop {{\mathop{\text{arg min}}\nolimits} }\limits_{{\boldsymbol{T}} \in {\mathop{\rm SE}\nolimits} \left( 3 \right)} {\mkern 1mu} {\cal L}({\boldsymbol{T}}\left| {{{\boldsymbol{I}}_q},{\mathcal{I}}, {\cal G},\boldsymbol{K},{\boldsymbol{T}_{ref}}, {\boldsymbol{T}_{c_l}}} \right.),
\end{equation}
where multiple rendered images is given by \(\mathcal{I} = \{\boldsymbol{I}_{ref}, \boldsymbol{I}_{c_1}, \boldsymbol{I}_{c_2}, \dots, \boldsymbol{I}_{c_L} \}\), with \(\boldsymbol{I}_{ref}\) denoting the reference image and \(\boldsymbol{I}_{c_l}\) representing the candidate image rendered from 3DGS corresponding to the initial pose \(\boldsymbol{T}\) and hypothetical poses \(\boldsymbol{T}_{c_l}\), respectively.

\subsection{Multi-view Consistency Strategy}\label{3.2}

Previous methods \cite{mcloc}, \cite{Transactions_uav_refinement_3dgs_2025} have demonstrated that features extracted from pre-trained CNN and Transformer backbones can be leveraged by constructing similarity measures for iterative pose refinement. However, the geometric relationships between the query image and candidate images remain insufficiently explored. In this section, we propose a multi-view consistency strategy that computes the refined pose between the relative pose of the query image and rendered images. Specifically, we first employ a pre-trained feature matching network to extract 2D-2D correspondences between these images. We then introduce a matching-based multi-view consistency procedure that enforces consistency among the refined pose, the relative pose, and the initial pose, guided by the modeled camera motion.

\subsubsection{2D-2D Matching via Pre-trained Model}
For the set of images, the proposed framework employs SuperPoint \cite{superpoint} to extract interest point locations and descriptors from each image. We denote the query image as $\boldsymbol{I}_q$ and the rendered images as $\mathcal{I} = \{\boldsymbol{I}_{ref}, \boldsymbol{I}_{c_1}, \dots, \boldsymbol{I}_{c_L}\}$. For each image, a feature extractor outputs interest point locations filtered by a detection threshold $\tau_{\text{detection}}$:
\begin{align}
\{\boldsymbol{u}_i\} &= F_\zeta(\boldsymbol{I}_q),\\
\{\boldsymbol{u}_j\} &= F_\zeta(\boldsymbol{I}_{s}),~\boldsymbol{I}_{s} \in \mathcal{I},
\end{align}
where images $\boldsymbol{I}_q$ and $\boldsymbol{I}_{s}$ have $M$ and $N$ interest points, indexed by $i = 1, \dots, M$ and $j = 1, \dots, N$, $F_\zeta(\cdot)$ denotes the interest point location extractor, $\boldsymbol{f}_i \in \mathbb{R}^3 $ and $ \boldsymbol{f}_j \in \mathbb{R}^3$ denote the homogeneous pixel coordinates, respectively.

Moreover, the corresponding interest point descriptors are extracted as:
\begin{align}
\{\boldsymbol{h}_i\} & = F_\Omega(\boldsymbol{I}_q) ,\\
\{\boldsymbol{h}_j\} & = F_\Omega(\boldsymbol{I}_s),~\boldsymbol{I}_r \in \mathcal{I},
\end{align}
where $F_\Omega(\cdot)$ is the descriptor extractor, $\boldsymbol{h}_i$ and $\boldsymbol{h}_j$ denote the normalized descriptors, respectively.
 
As for the matching stage, we employ LightGlue \cite{lightglue} to predict an assignment matrix $\boldsymbol{P} \in [0,1]^{M \times N}$ representing the matchability between the query and rendered image features. Each element $p_{ij} \in \boldsymbol{P}$ indicates the normalized pairwise matchability between the $i$-th feature from $\boldsymbol{I}_q$ and the $j$-th feature from $\boldsymbol{I}_s$. Feature correspondences are obtained by:
\begin{align}
\{\boldsymbol{u}_{i}^{\prime} \leftrightarrow \boldsymbol{u}_{j}^{\prime} \mid p_{ij} > \tau_{\text{matching}}\},
\end{align}
where $\tau_{\text{matching}}$ is a threshold applied to filter out low-confidence feature pairs.

\subsubsection{Matching-Based Multi-view Consistency}

In our approach, the refined pose is derived from the estimated relative poses between the query image and the rendered images. Based on the established 2D-2D feature correspondences, we estimate the relative pose between the query image \({{\boldsymbol{I}}_{q}}\) and the rendered image \(\boldsymbol{I}_{s} \in \mathcal{I}\) as follows:
\begin{equation}
\label{eq:relative}
{\boldsymbol{u}_j^{\prime\rm{T}}}{\boldsymbol{K}^{-{\rm{T}}}} s[{\boldsymbol{t}_{sq}} ]^{\wedge} {\boldsymbol{R}_{sq}}{\boldsymbol{K}^{-1}}{\boldsymbol{u}_i^{\prime}} = 0,
\end{equation}
where $\boldsymbol{K}$ represents the camera intrinsic matrix, $\boldsymbol{R}_{sq} \in \mathrm{SO}(3)$ denotes the relative rotation matrix, $s\boldsymbol{t}_{sq} \in \mathbb{R}^3$ is the scaled relative translation vector, and $[\boldsymbol{t}_{sq}]^{\wedge}$ denotes the skew-symmetric matrix of the normalized translation vector $\boldsymbol{t}_{sq}$. Since Eq.~(\ref{eq:relative}) is equal to zero, it inherently introduces scale ambiguity, preventing the direct computation of the absolute translation scale. Here, $s$ denotes the unknown scale factor that relates the true translation to the normalized vector $\boldsymbol{t}_{sq}$. To ensure robustness against outlier correspondences, the RANSAC algorithm \cite{ransac} is employed during the relative pose estimation process. Therefore, the estimated relative poses between \({{\boldsymbol{I}}_{q}}\) and \(\mathcal{I} = \{\boldsymbol{I}_{ref}, \boldsymbol{I}_{c_1}, \dots, \boldsymbol{I}_{c_L} \}\) are defined up to an unknown scale, and can be expressed as follows:
\begin{align}
\label{eq:relative-qr-scale}
{{\boldsymbol{T}}_{qr}} &= 
\begin{bmatrix} 
    {\boldsymbol{R}_{qr}} & s_{qr}{\boldsymbol{t}_{qr}} \\ 
    0 & 1 
\end{bmatrix}, \\
\label{eq:relative-qc-scale}
{{\boldsymbol{T}}_{q{c_l}}} &= 
\begin{bmatrix} 
    {\boldsymbol{R}_{q{c_l}}} & s_{q{c_l}}{\boldsymbol{t}_{q{c_l}}} \\ 
    0 & 1 
\end{bmatrix},
\end{align}
where \({{\boldsymbol{T}}_{qr}}\) represents the relative pose between the reference image and the query image, and \({{\boldsymbol{T}}_{q{c_l}}}\) represents the relative pose between the candidate image and the query image, $l = 1, \dots, L$ denotes the index of each candidate image in the set.

To address the scale ambiguity, we establish the pose motion relationship across multiple views, as illustrated in the second step of Fig.~\ref{fig:illustrated_part}, which can be expressed by:
\begin{align}
\label{eq:relative-qr}
{\boldsymbol{\hat T}} &= {{{\boldsymbol{T}}_{qr}}{\boldsymbol{T}_{ref}}}, \\
\label{eq:relative-qc}
{\boldsymbol{\hat T}} &= {{\boldsymbol{T}}_{q{c_l}}}{{\boldsymbol{T}}_{{c_l}}}.
\end{align}

By substituting Eq. (\ref{eq:relative-qr-scale}) and Eq. (\ref{eq:relative-qc-scale}) into Eq. (\ref{eq:relative-qr}) and Eq. (\ref{eq:relative-qc}), we derive the formulation for predicting the scale factors among the relative rotations, translations and the absolute rotations, translations, which can be expressed as follows:
\begin{align}
    \boldsymbol{R}_{qr}&=\boldsymbol{R}_{q{c_l}}\boldsymbol{R}_{c_l}\boldsymbol{R}_{ref}^{-1}, \\
    s_{qr}{\boldsymbol{t}_{qr}} &={\boldsymbol{R}_{q{c_l}}} + s_{q{c_l}}{\boldsymbol{t}_{q{c_l}}}-\boldsymbol{R}_{qr}\boldsymbol{t}_{ref}.
\end{align}

After all the pairwise transformations are initialized, when it comes to multiple candidate images, \(s_{qr}\) can be estimated with \(L\) possible values due to the numerous \(s_{q{c_l}}\), leading to different precise predictions for each candidate image. To ensure multi-view consistency among various solutions, we minimize the discrepancy based on two-view relative transformations and camera pose motion relationship, and thereby simultaneously obtain the calculated values, which can be formulated as:
\begin{align}
    \label{eq:coarse-pose}
    & \mathop{\arg \min}\limits_{{\boldsymbol{\hat R}},{\boldsymbol{\hat t}},{s_{qr}},{s_{q{c_l}}}} \frac{1}{{L + 1}} \Bigg(
    \left\| {\log ({{\boldsymbol{\hat R}}^{ - 1}}{{\boldsymbol{R}}_{qr}}{{\boldsymbol{R}}_{ref}})}  \right\|_2^2  \notag  \\
    & \quad + \left\| {{{\boldsymbol{\hat R}}^{ - 1}}\left({{\boldsymbol{R}}_{qr}}{{\boldsymbol{t}}_{ref}}+{{s}_{qr}}{{\boldsymbol{t}}_{qr}}-{\boldsymbol{\hat t}}\right)} \right\|_2^2  \notag \\
    & \quad + \sum\limits_{l = 1}^L {\left\| {\log ({{\boldsymbol{\hat R}}^{ - 1}}{{\boldsymbol{R}}_{q{c_l}}}{{\boldsymbol{R}}_{{c_l}}})} \right\|_2^2 }  \notag  \\
    & \quad + \sum\limits_{l = 1}^L {\left\| {{{\boldsymbol{\hat R}}^{ - 1}}\left({{\boldsymbol{R}}_{q{c_l}}}{{\boldsymbol{t}}_{c_l}}+{{s}_{q{c_l}}}{{\boldsymbol{t}}_{q{c_l}}}-{\boldsymbol{\hat t}}\right)} \right\|_2^2 } \Bigg).
\end{align}

\subsection{Epipolar Geometric Constrained Optimization}\label{3.3}

While learning-based feature matching models combined with standard outlier rejection provide reliable correspondences, residual mismatches often persist. Such mismatches often arise from intrinsic ambiguities in textureless regions, repetitive patterns, or inaccuracies in geometric reconstruction and surface texture. Mitigating these mismatches is essential for preserving multi-view consistency and enhancing pose refinement accuracy. To address these limitations, we further enhance the robustness and accuracy of camera pose optimization by incorporating epipolar geometry constraints.

With the estimated camera pose obtained by Eq. (\ref{eq:coarse-pose}), and according to the correspondences between the 2D point \({{\boldsymbol{u}}_i}\) of the image \({{\boldsymbol{I}}_{q}}\) and the 2D point \({{\boldsymbol{u}}_j^{\prime}}\) of the image \({{\boldsymbol{I}}_{s}}\), the \({{\boldsymbol{u}}_i^{\prime}}\) can be projected onto the unique epipolar line of \({{\boldsymbol{I}}_{s}}\). The coefficients of epipolar line on the \({{\boldsymbol{I}}_{s}}\) can be defined as:
\begin{align}
{\boldsymbol{p}_{sj}} 
& = {\boldsymbol{K}^{-\rm{T}}} [{\boldsymbol{t}_{sq}} ]^{\wedge} {\boldsymbol{R}_{sq}}{\boldsymbol{K}^{-1}}{\boldsymbol{u}_i^{\prime}} \notag \\ 
& = \begin{bmatrix} {p}_{{sj}_x} \ {p}_{{sj}_y} \ {p}_{{sj}_c} \end{bmatrix}^{\rm T},
\end{align}
where \({\boldsymbol{t}_{sq}}\) denotes the translation vector with scale factor.

After constructing the epipolar line equation, the distance between the pixel location \(\boldsymbol{u}_j^{\prime}\) and its corresponding epipolar line \(\boldsymbol{p}_{sj}\) can be quantified as:
\begin{equation}
{d}_{ij} = \frac{{\boldsymbol{u}}_j^{\prime{T}}{{\boldsymbol{p}_{sj}}}}{{\sqrt {{{{p}_{{sj}_x}}^2} + {{p}_{{sj}_y}^2}} }}.
\end{equation}

The camera pose is optimized through a nonlinear process that minimizes the sum of epipolar line distances over all matched points, as illustrated in the third step of Fig.~\ref{fig:illustrated_part}. However, the presence of outliers can compromise the robustness of the standard least squares approach. To address this issue, we propose a robust objective function that leverages the residual standard deviation to suppress the influence of outliers:
\begin{equation}
{\boldsymbol{\hat T}^*} = \mathop{\arg}\mathop{\min}\limits_{\boldsymbol{\hat T}}\left\{ { F(\boldsymbol{\hat T}) }\right\} ,
\end{equation}
where \(F(\boldsymbol{\hat T})\) can be derived as:
\begin{align}
\frac{1}{2} \left[ \sum\limits_{j_{ref}=1}^{N_{ref}} \sum\limits_{i=1}^{M}
\rho\left(\frac{d_{i j_{ref}}}{\hat{\sigma}_{i j_{ref}}}\right)
+ \frac{1}{L} \sum\limits_{l=1}^L  \sum\limits_{j_{c_{l}}=1}^{N_{c_{l}}} \sum\limits_{i=1}^{M} 
\rho\left(\frac{d_{i j_{c_{l}}}}{\hat{\sigma}_{i j_{c_{l}}}}\right) \right].
\end{align}

\begin{figure}[!t]
    \centering
    \subfloat[5-tuples images]{\includegraphics[width=1\linewidth]{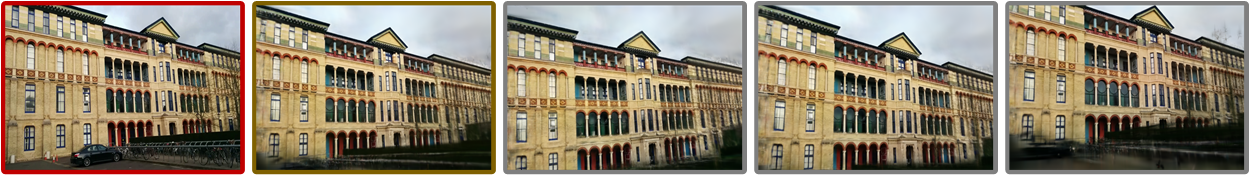}}
    \\
    \vspace{-0.8em}\subfloat[Distance \({r_{ij}}\)]{\includegraphics[width=1\linewidth, trim=0 5 3 3, clip]{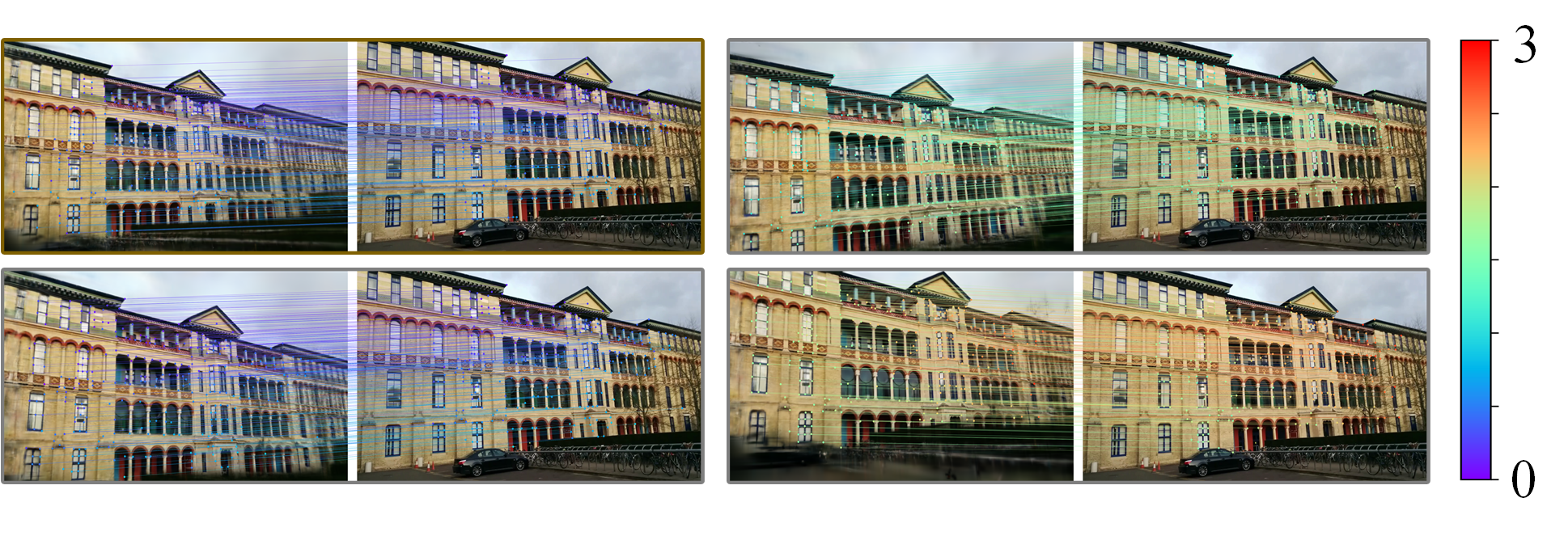}}
    \caption{Illustration of scaled distance \({r_{ij}}\). (a) When $L=3$, the input consists of a 5-tuple: one query image, one reference image, and three candidate images. (b) The scaled distance from each detected feature point to its corresponding epipolar line, shown on the \textit{Hospital}.}
    \label{fig:visualization_EGC}
\end{figure}

Here, we use the Huber function to penalize residuals. Let \({r_{ij} = d_{ij}/\hat{\sigma}_{ij}}\). Fig.~\ref{fig:visualization_EGC} illustrates the scaled distance \(r_{ij}\) between the matched points and their corresponding epipolar lines. Then we have:
\begin{align}
\rho \left(r_{ij}\right) &= 
\begin{cases}
    {{(r_{ij})}^2}, & \text{if } {\left| {r_{ij}} \right|} \leq 1 \\
    2 (r_{ij})^{\frac{1}{2}}-1, & \text{otherwise},
\end{cases} \\
\hat{\sigma}_{ij} &= \sqrt{\frac{1}{(m-1)(n-1)} \sum_{j=1}^{m} \sum_{i=1}^{n} d_{i j}^{2}}.
\end{align}

For precise pose optimization, we adopt the Levenberg--Marquardt (LM) algorithm \cite{lm} to approximate the solution of the nonlinear least squares problem in our refinement framework. Directly optimizing in the special Euclidean group $\mathrm{SE}(3)$ is challenging due to its six degrees of freedom being represented by redundant parameters in rotation matrices and translation vectors. To address this, we parameterize the camera pose in the associated Lie algebra $\mathfrak{se}(3)$, which provides a minimal and locally Euclidean representation suitable for gradient-based optimization. Specifically, the rotation and translation vectors are expressed as \(\mathfrak{se}(3)= \left\{ {\bf{\xi }} = {\left[ {\begin{array}{*{2}{@{\hskip 2pt}c@{\hskip 2pt}}}{\boldsymbol{\rho }}{\hskip 6pt} {\boldsymbol\phi} \end{array}} \right]^{\rm{T}}} \in {{\mathbb R}^6}, {\boldsymbol\rho} \in {{\mathbb R}^3},  {\boldsymbol\phi} \in \mathfrak{so}(3) \right\} \). Let \({\boldsymbol\phi} = \theta  \boldsymbol{a} \) and \({\boldsymbol t} = {\boldsymbol{J} \boldsymbol{\rho}} \), where \(\boldsymbol{J}\) is computed as:
\begin{align}
\boldsymbol{J} = \frac{\sin \theta}{\theta} \boldsymbol{I} + \left(1 - \frac{\sin \theta}{\theta}\right) \boldsymbol{a} \boldsymbol{a}^{\rm{T}} + \frac{1 - \cos \theta}{\theta} \boldsymbol{a}^{\wedge},
\end{align}
where \(\theta\) denotes the magnitude of \(\boldsymbol{\phi}\), and \(\boldsymbol{a}\) represents the corresponding unit direction vector. Ultimately, the trajectory converges after several iterations of the optimization process.

\section{Experiments}\label{4}

In this section, we first describe the datasets, metrics, and implementation details. Then, we compare our method against baseline methods. Furthermore, we conduct ablation studies to validate the effectiveness of each component in our method. Finally, we analyze failure cases and limitations of our approach.

\subsection{Datasets and Metrics}\label{4.1}
To evaluate the performance of our pose refinement framework, we conduct experiments on the 7-Scenes and the Cambridge Landmarks datasets, which are commonly used in the visual localization task. 

\subsubsection{7-Scenes Dataset}
The Microsoft 7-Scenes dataset \cite{7scenes}\cite{7scenes_sfm_gt} is a widely used benchmark for indoor visual localization. 7-Scenes includes scenes such as \textit{Chess, Fire, Heads, Office, Pumpkin, RedKitchen, and Stairs}, with areas ranging from 1 \({m^2}\) to 18 \({m^2}\). This dataset presents significant challenges for visual relocalization due to repetitive structures, textureless regions, and motion blur, all of which introduce unique difficulties during the pose estimation process.

\subsubsection{Cambridge Landmarks Dataset}
The Cambridge Landmarks dataset \cite{posenet_cambridge_2015} is a large-scale outdoor benchmark that covers urban scenes around Cambridge University, including \textit{King, Hospital, Shop, and Church}, with scene sizes ranging from 875 \({m^2}\) to 5600 \({m^2}\). The Cambridge dataset is particularly challenging due to varying lighting conditions and numerous moving distractors, such as pedestrians and vehicles, present in different sequences of recorded data.

\begin{table*}[!t]
    \captionsetup{justification=centering, labelsep=none} 
    \caption{Quantitative results on the indoor 7-Scenes Dataset. \\We report median translation error ${\rm{e}}({\boldsymbol{\hat t}})(cm)$ and median rotation error ${\rm{e}}({\boldsymbol{\hat R}})(^\circ)$. \label{tab:7Scenes_output}}
    \centering
    \tabcolsep=0.010\linewidth
    \begin{tabular}{@{\hspace{-1pt}}clcccccccccccccc|cc}
        \toprule
        \multirow{2}{*}{} & \multicolumn{1}{l}{\multirow{2}{*}{Methods}} & \multicolumn{2}{c}{Chess} & \multicolumn{2}{c}{Fire} & \multicolumn{2}{c}{Heads} & \multicolumn{2}{c}{Office} & \multicolumn{2}{c}{Pumpkin} & \multicolumn{2}{c}{RedKitchen} & \multicolumn{2}{c|}{Stairs} & \multicolumn{2}{c}{Average $\downarrow$ } \\
        & & ${\rm{e}}({\boldsymbol{\hat t}})$ & ${\rm{e}}({\boldsymbol{\hat R}})$ & ${\rm{e}}({\boldsymbol{\hat t}})$ & ${\rm{e}}({\boldsymbol{\hat R}})$ & ${\rm{e}}({\boldsymbol{\hat t}})$ & ${\rm{e}}({\boldsymbol{\hat R}})$ & ${\rm{e}}({\boldsymbol{\hat t}})$ & ${\rm{e}}({\boldsymbol{\hat R}})$ & ${\rm{e}}({\boldsymbol{\hat t}})$ & ${\rm{e}}({\boldsymbol{\hat R}})$ & ${\rm{e}}({\boldsymbol{\hat t}})$ & ${\rm{e}}({\boldsymbol{\hat R}})$ & ${\rm{e}}({\boldsymbol{\hat t}})$ & ${\rm{e}}({\boldsymbol{\hat R}})$ & ${\rm{e}}({\boldsymbol{\hat t}})$ & ${\rm{e}}({\boldsymbol{\hat R}})$ \\
        \hline
        \multirow{4}{*}{\rotatebox{90}{APR}} & Ms-Transformer \cite{ms-transformer_2021}  & 11 & 4.66 & 24  & 9.6 & 14 & 12.19 & 17 & 5.66 & 18 & 4.44 & 17 & 5.94 & 26 & 8.45 & 31.75 & 7.27\\
        & DFNet \cite{dfnet}    & 3.3 & 1.12 & 6  & 2.3 & 4 & 2.29 & 6 & 1.54 & 7 & 1.92 & 7 & 1.74 & 12 & 2.63 & 6.47 & 1.93\\
        & LENS \cite{lens}     & 3 & 1.3 & 10  & 3.7 & 7 & 5.8 & 7 & 1.9 & 8 & 2.2 & 9 & 2.2 & 14 & 3.6 & 8.28 & 2.96\\
        & Marepo \cite{cs-marepo}     & 2.1 & 1.24 & 2.3 & 1.39 & 1.8 & 2.03 & 2.8 & 1.26 & 3.5 & 1.48 & 4.2 & 1.71 & 5.6 & 1.67 & 3.19 & 1.54\\ 
        \hline
        \multirow{7}{*}{\rotatebox{90}{Pose Refiners}} & FQN \cite{FQN}   & 4 & 1.3 & 10  & 3.0 & 4 & 2.4 & 10 & 3.0 & 9 & 2.4 & 16 & 4.4 & 140 & 34.7 & 25.57 & 7.31\\
        & CrossFire \cite{crossfire}    & 1 & 0.4 & 5  & 1.9 & 3 & 2.3 & 5 & 1.6 & 3 & 0.8 & 2 & 0.8 & 12 & 1.9 & 4.43 & 1.39\\
        & DFNet+NeFeS \cite{NeFeS50}   & 2 & 0.57 & 2  & 0.74 & 2 & 1.28 & 2 & 0.56 & 2 & 0.55 & 2 & 0.57 & 5 & 1.28 & 2.42 & 0.79\\
        & DFNet+HR-APR \cite{cs-hr-apr}    & 2 & 0.55 & 2  & 0.75 & 2 & 1.45 & 2 & 0.64 & 2 & 0.62 & 2 & 0.67 & 5 & 1.30 & 2.42 & 0.85\\
        & MCLoc \cite{mcloc}    & 2 & 0.8 & 3  & 1.4 & 3 & 1.3 & 4 & 1.3 & 5 & 1.6 & 6 & 1.6 & 6 & 2.0 & 4.14 & 1.42 \\
        & DFNet+GS-SMC (ours)    & \underline{0.65} & \underline{0.21} & \underline{0.97}  & \underline{0.38} & \underline{0.71} & \underline{0.44} & \underline{1.5} & \underline{0.40} & \underline{1.5} & \underline{0.31} & \underline{1.1} & \underline{0.28} & \underline{2.9} & \underline{0.81} & \underline{1.33} & \underline{0.40} \\
        & Marepo+GS-SMC (ours)     & \textbf{0.64} & \textbf{0.20} & \textbf{0.88}  & \textbf{0.36} & \textbf{0.70} & \textbf{0.43} & \textbf{1.4} & \textbf{0.37} & \textbf{1.4} & \textbf{0.28} & \textbf{1.0} & \textbf{0.26} & \textbf{1.9} & \textbf{0.54} & \textbf{1.13} & \textbf{0.34}\\
        \bottomrule
    \end{tabular}
\end{table*}
    
\begin{figure*}[!t]
    \centering
    \captionsetup[subfloat]{labelformat=empty}
    \subfloat[]{\includegraphics[width=0.24\linewidth]{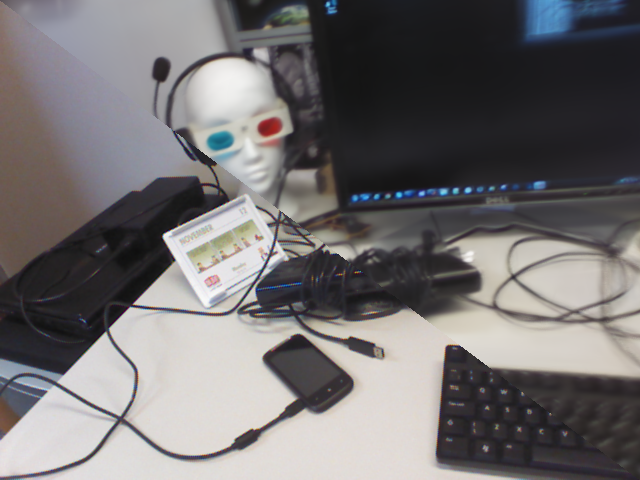}}
    \hfill
    \subfloat[]{\includegraphics[width=0.24\linewidth]{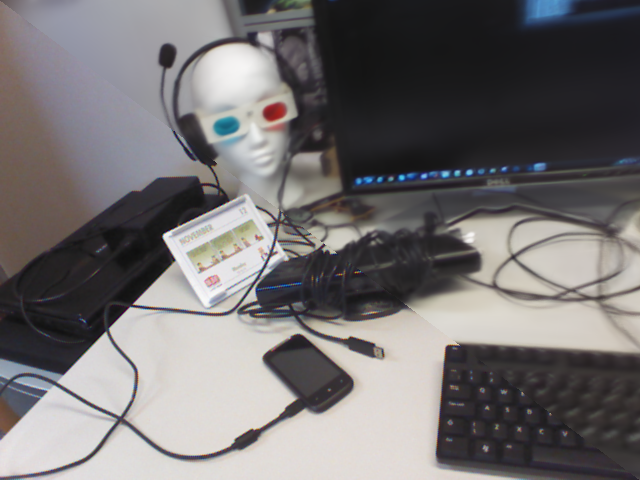}}
    \hfill
    \subfloat[]{\includegraphics[width=0.24\linewidth]{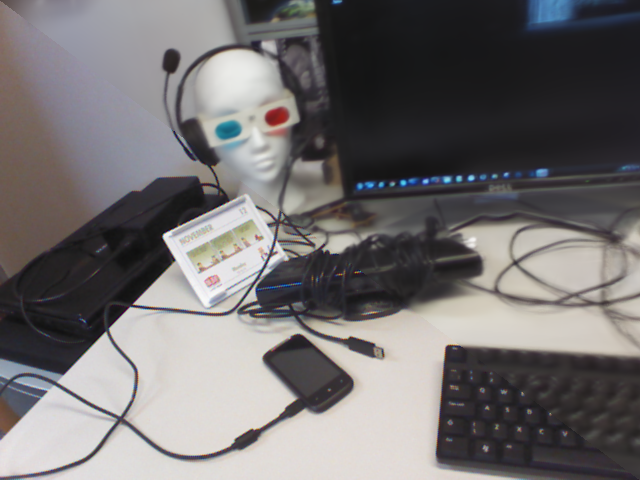}}
    \hfill
    \subfloat[]{\includegraphics[width=0.24\linewidth]{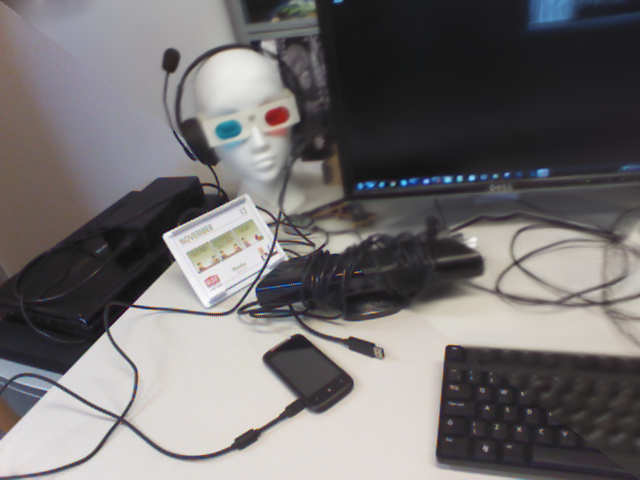}}\\
    \captionsetup[subfloat]{labelformat=empty}
    \vspace{-0.7cm} 
    \subfloat[]{\includegraphics[width=0.24\linewidth]{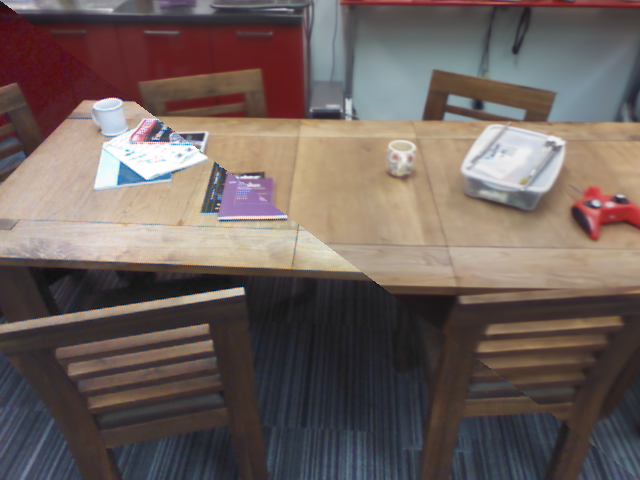}}
    \hfill
    \subfloat[]{\includegraphics[width=0.24\linewidth]{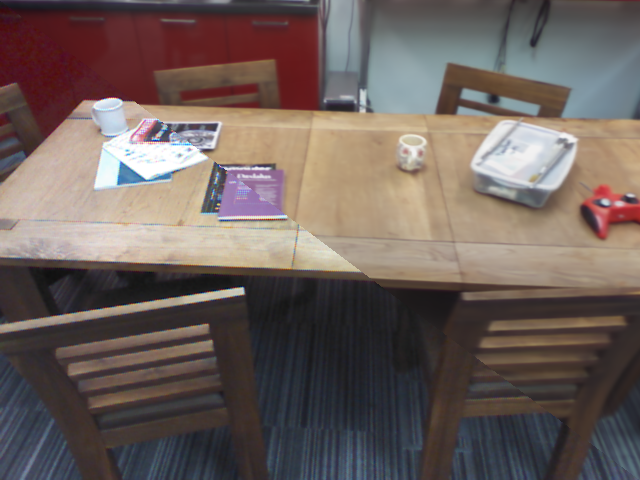}}
    \hfill
    \subfloat[]{\includegraphics[width=0.24\linewidth]{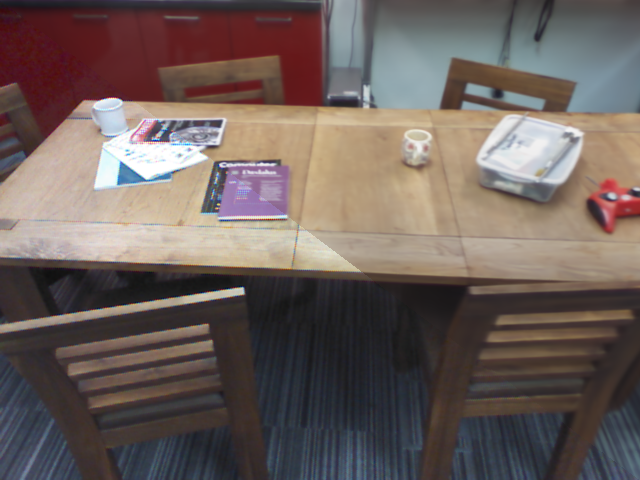}}
    \hfill
    \subfloat[]{\includegraphics[width=0.24\linewidth]{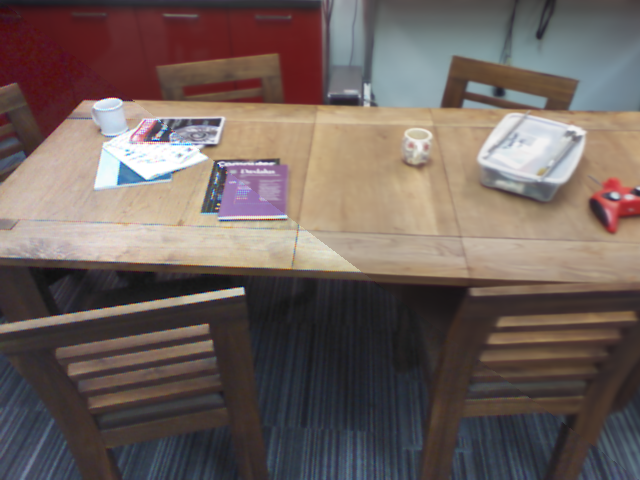}}\\
    \captionsetup[subfloat]{labelformat=parens}
    \setcounter{subfigure}{0}
    \vspace{-0.7cm} 
    \subfloat[DFNet \cite{dfnet} \label{fig:dfnet}]{\includegraphics[width=0.24\linewidth]{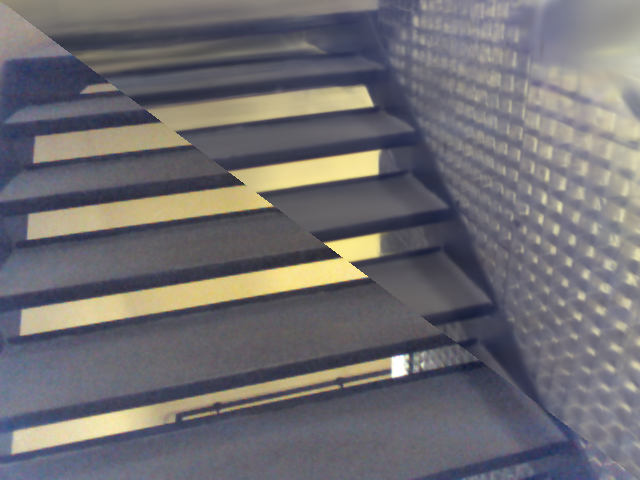}}
    \hfill
    \subfloat[Marepo \cite{cs-marepo} \label{fig:marepo}]{\includegraphics[width=0.24\linewidth]{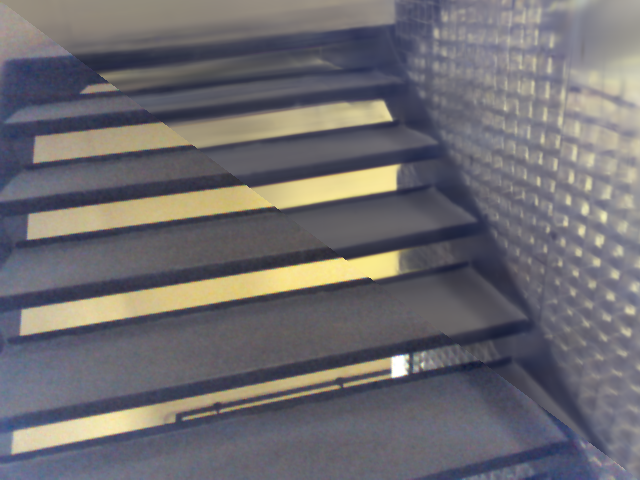}}
    \hfill
    \subfloat[D.+GS-SMC (ours) \label{fig:dfnet+ours}]{\includegraphics[width=0.24\linewidth]{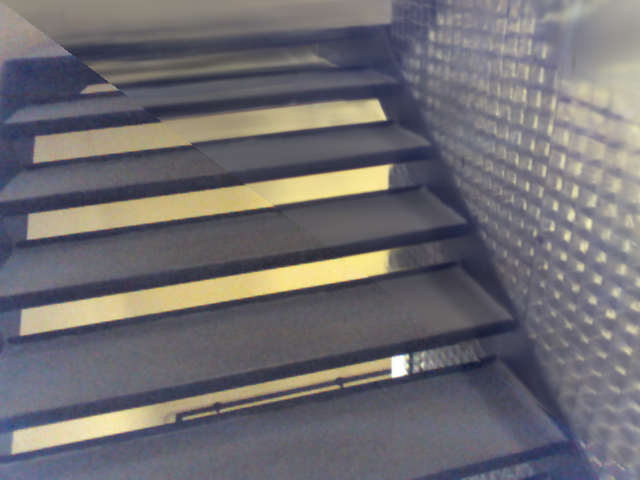}}
    \hfill
    \subfloat[M.+GS-SMC (ours) \label{fig:mar+ours}]{\includegraphics[width=0.24\linewidth]{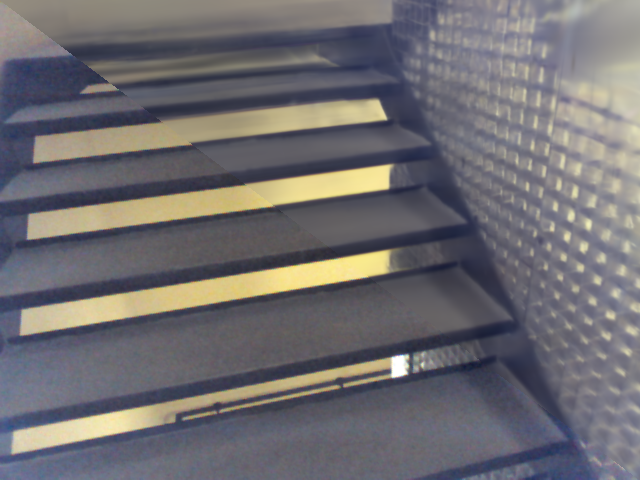}}
    \caption{Visualization of GS-SMC improves pose prediction for DFNet and Marepo on the 7-Scenes dataset. Each subfigure is split diagonally, with the bottom left displaying the ground truth query image and the top right showing the rendered image with the estimated/refined pose. (1st row: the \textit{Heads} scene, 2nd row: the \textit{RedKitchen} scene, 3rd row: the \textit{Stairs} scene.)  \label{fig:visualization_7scenes}}
\end{figure*}

\subsubsection{Metrics}
To evaluate the accuracy of the pose estimation results, we use the median translation and rotation error as evaluation metrics following \cite{posenet_cambridge_2015}. The translation error \({\rm{e}}({\boldsymbol{\hat t}})\) and the rotation error \({\rm{e}}({\boldsymbol{\hat R}})\) are calculated by:

\begin{align}
    {{\rm{e}}({\boldsymbol{\hat t}})} & =\left\|\boldsymbol{\hat t}-\boldsymbol{t}_{\rm{gt}}\right\|_{2}\\
    {{\rm{e}}({\boldsymbol{\hat R}})} & = \arccos \left( (\operatorname{trace}(\hat{\boldsymbol R}^{\mathrm T}\boldsymbol R_{\mathrm{gt}})-1)/2 \right)
\end{align}
where \({\boldsymbol{\hat R}}\) and \({\boldsymbol{\hat t}}\) represent the refinement rotation and translation of methods and \({\boldsymbol{R}}_{\rm{gt}}\) and \({\boldsymbol{t}}_{\rm{gt}}\) denote the ground-truth rotation and translation. And \(trace\left(\cdot\right)\) denotes the trace of the matrix.

We compete against the following baselines:

\paragraph{APR Methods}
We compare our method to against Absolute Pose Regression (APR) techniques, which train end-to-end networks to directly regress the camera pose from the query image, including Ms-Transformer \cite{ms-transformer_2021}, DFNet \cite{dfnet}, LENS \cite{lens}, and Marepo \cite{cs-marepo}.

\paragraph{Camera Pose Refinement}
We evaluate our method against pose refiners such as FQN \cite{FQN}, CrossFire \cite{crossfire}, NeFeS \cite{NeFeS50}, and HR-APR \cite{cs-hr-apr}, which rely on dedicated neural networks with NeRF scene representation. In contrast, MCLoc \cite{mcloc} leverages 3DGS and generic pre-trained features to compute feature-level similarity and refines the camera pose through Monte Carlo sampling.

\begin{remark}
We report median translation error ${\rm{e}}({\boldsymbol{\hat t}})$ and median rotation error ${\rm{e}}({\boldsymbol{\hat R}})$. In Table~\ref{tab:7Scenes_output}--\ref{tab:ablation_impact}, the best results for each category are in bold, and the second-best results are underlined. Results that are both underlined and italicized correspond to the results reported in Table~\ref{tab:Cambridge_output}. "TS" denotes per-scene training of dedicated neural networks, and "-" indicates that no such training is required. ``DFNet/Marepo + GS-SMC" represents that our proposed refinement process integrates the result of DFNet \cite{dfnet} and Marepo \cite{cs-marepo} as the initial pose. ``+w/o. multi-view" indicates the ablation study without a multi-view consistency strategy. ``+w/o. EGC" indicates the ablation study without epipolar geometry constraints.
\end{remark}

\subsection{Implementation Details}\label{4.2}
For the 7-Scenes dataset, as demonstrated in \cite{NeFeS50}, \cite{cs-marepo}, the SfM ground truth (GT) \cite{7scenes_sfm_gt} is capable of rendering more detailed scene structures and provides enhanced geometric accuracy compared to dSLAM GT for the 7-Scenes dataset. Additionally, for the Cambridge Landmarks dataset, we employ the GT provided by \cite{posenet_cambridge_2015}. 

In the experimental implementation, we employ the official pre-trained SuperPoint \cite{superpoint} and LightGlue \cite{lightglue} models for 2D-2D correspondences, without fine-tuning. Our method refines the pose iteratively, with each iteration computing the pose by estimating the relative poses between the query image and the rendered images. We set the number of iterations to six for the 7-Scenes dataset, while perturbing the initial pose with Gaussian noise, with a standard deviation of 4 cm for translation and 0.01$^\circ$ for rotation, to render candidate images. For the Cambridge Landmarks dataset, the refinement process is terminated after four steps, with the initial pose perturbed by Gaussian noise with standard deviations of 4 m for translation and 2$^\circ$ for rotation.

\begin{figure}[!t]
    \centering
    \subfloat[\label{7scenes_iter_median_trans}]{\includegraphics[width=0.49\linewidth, trim=2 2 15 10, clip]{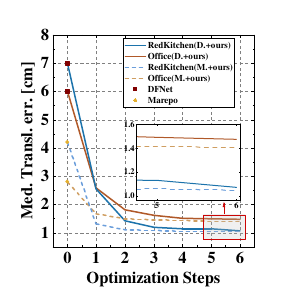}}
    \hspace{0.001\linewidth}
    \subfloat[\label{7scenes_iter_median_rotation}]{\includegraphics[width=0.49\linewidth, trim=2 2 15 10, clip]{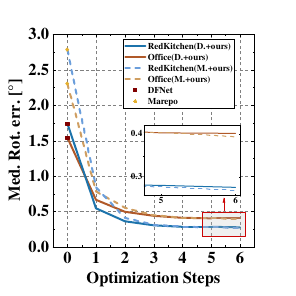}}
    \caption{Optimization trajectory on the 7-Scenes dataset. We present the variation in median error with respect to the number of optimization iterations for DFNet+GS-SMC (ours) and Marepo+GS-SMC (ours) across two scenes. \label{fig:visualization_7scenes_iter}}
\end{figure}

\begin{table*}[!t]
    \captionsetup{justification=centering, labelsep=none} 
    \centering
    \caption{Comparison results on the Cambridge Landmarks dataset.  \label{tab:Cambridge_output}}
    \centering
    \tabcolsep=0.016\linewidth
    \begin{tabular}{@{\hspace{-1pt}}clccccccccc|cc}
        \toprule
        \multirow{2}{*}{} &\multicolumn{1}{l}{\multirow{2}{*}{Methods}} & \multirow{2}{*}{} & \multicolumn{2}{c}{Kings} & \multicolumn{2}{c}{Hospital} & \multicolumn{2}{c}{Shop} & \multicolumn{2}{c|}{Church} & \multicolumn{2}{c}{ Average $\downarrow$}\\
        & & & ${\rm{e}}({\boldsymbol{\hat t}}) $ & ${\rm{e}}({\boldsymbol{\hat R}}) $ & ${\rm{e}}({\boldsymbol{\hat t}})$ & ${\rm{e}}({\boldsymbol{\hat R}})$ &  ${\rm{e}}({\boldsymbol{\hat t}})$ & ${\rm{e}}({\boldsymbol{\hat R}})$ & ${\rm{e}}({\boldsymbol{\hat t}})$ & ${\rm{e}}({\boldsymbol{\hat R}})$ & ${\rm{e}}({\boldsymbol{\hat t}})$ & ${\rm{e}}({\boldsymbol{\hat R}})$\\  
        \hline
        \multirow{3}{*}{\rotatebox{90}{APR}} & Ms-Transformer \cite{ms-transformer_2021} & TS & 0.83 m & 1.47$^\circ$ & 1.81 m & 2.39$^\circ$ & 0.86 m & 3.07$^\circ$ & 1.62 m & 3.99$^\circ$ &1.28 m &2.73$^\circ$ \\
        & DFNet \cite{dfnet}   & TS  & 0.73 m & 2.37$^\circ$ & 2.0 m & 2.98$^\circ$ & 0.67 m & 2.21$^\circ$ & 1.37 m & 4.03$^\circ$ &1.19 m &2.90$^\circ$ \\
        & LENS \cite{lens}    & TS  & 0.33 m & 0.5$^\circ$ & 0.44 m & 0.9$^\circ$ & 0.25 m & 1.6$^\circ$ & 0.53 m & 1.6$^\circ$ & 0.39 m & 1.15$^\circ$ \\
        \hline 
        \multirow{6}{*}{\rotatebox{90}{Pose Refiners}} & FQN \cite{FQN}   & TS  & 0.28 m & 0.4$^\circ$ & 0.54 m & 0.8$^\circ$ & 0.13 m & 0.6$^\circ$ & 0.58 m & 2.0$^\circ$ &0.38 m &0.95$^\circ$ \\
        & CrossFire \cite{crossfire}   & TS  & 0.47 m & 0.7$^\circ$ & 0.43 m & 0.7$^\circ$ & 0.2 m & 1.2$^\circ$ & 0.39 m & 1.4$^\circ$ &0.37 m &1.0$^\circ$ \\
        & DFNet+NeFeS \cite{NeFeS50}   & TS  & 0.37 m & 0.54$^\circ$ & 0.52 m & 0.88$^\circ$ & 0.15 m & 0.53$^\circ$ & 0.37 m & 1.14$^\circ$ &0.35 m & 0.77$^\circ$  \\
        & DFNet+HR-APR \cite{cs-hr-apr}   & TS  & 0.36 m & 0.58$^\circ$ & 0.53 m & 0.89$^\circ$ & 0.13 m & 0.51$^\circ$ & 0.38 m & 1.16$^\circ$ &0.35 m &0.79$^\circ$ \\
        & MCLoc \cite{mcloc}   & -  & 0.31 m & 0.42$^\circ$ & 0.39 m & 0.73$^\circ$ & 0.12 m & 0.45$^\circ$ & 0.26 m & 0.88$^\circ$ &0.27 m & 0.62$^\circ$ \\
        & DFNet+GS-SMC (ours)   & -  & \textbf{0.24} m & \textbf{0.26}$^\circ$ & \textbf{0.25} m & \textbf{0.39}$^\circ$ & \textbf{0.05} m & \textbf{0.24}$^\circ$ & \textbf{0.10} m & \textbf{0.28}$^\circ$ & \textbf{0.16} m & \textbf{0.29}$^\circ$ \\
        \bottomrule
    \end{tabular}
\end{table*}

\subsection{Results on the 7-Scenes}\label{4.3}

In this section, we evaluate our method on the 7-Scenes dataset, comparing it against four leading APR methods \cite{ms-transformer_2021}, \cite{dfnet}, \cite{lens}, \cite{cs-marepo} and five state-of-the-art pose refinement approaches \cite{FQN}, \cite{crossfire}, \cite{NeFeS50}, \cite{cs-hr-apr}, \cite{mcloc}.

As a pose refinement method, GS-SMC can be seamlessly integrated into existing pose estimation pipelines to enhance their performance. In this experiment, we use the estimated poses from DFNet \cite{dfnet} and Marepo \cite{cs-marepo} as initial inputs to evaluate our approach. As shown in Table~\ref{tab:7Scenes_output}, without requiring task-specific training or additional fine-tuning, Marepo+GS-SMC achieves superior localization performance, outperforming all compared APR and pose refinement methods, with DFNet+GS-SMC achieving the second-best results. Specifically, compared to using Marepo \cite{cs-marepo} alone as the initial pose, GS-SMC further reduces the median translation and rotation errors by 2.06 cm and 1.2$^\circ$, demonstrating its effectiveness as a pose refinement module. Moreover, compared to state-of-the-art refinement techniques, Marepo+GS-SMC reduces the average median translation error and rotation error by 1.29 m and 0.45$^\circ$ for NeFeS \cite{NeFeS50}, 1.29 m and 0.51$^\circ$ for HR-APR \cite{cs-hr-apr}, and 3.01 m and 1.08$^\circ$ for MCLoc \cite{mcloc}. 

Furthermore, as shown in Fig.\ref{fig:visualization_7scenes}, the rendered images using our results are closer to the query images, even in challenging scenes with repetitive textures, significantly improving the performance of both DFNet and Marepo. Notably, integrating our method with Marepo achieves better results, as shown in Fig.\ref{fig:visualization_7scenes}\subref{fig:dfnet+ours}, Fig.\ref{fig:visualization_7scenes}\subref{fig:mar+ours}, and Table~\ref{tab:7Scenes_output}. While a better initial pose further improves the refinement results of our method, this improvement is limited in some cases.

Additionally, on the 7-Scenes dataset, we set the optimization to six refinement steps, although satisfactory results are often achieved much earlier, as shown in Fig.~\ref{fig:visualization_7scenes_iter}. The optimization trajectory reveals that more accurate initial pose estimation improves the refinement process and reduces the number of iterations required. Overall, we conclude that our method significantly exceeds the performance of state-of-the-art refinement methods in terms of accuracy.

\subsection{Results on the Cambridge Landmarks}\label{4.4}

\begin{figure*}[!t]
\centering
\captionsetup[subfloat]{labelformat=empty}
\subfloat[]{\includegraphics[width=0.24\linewidth]{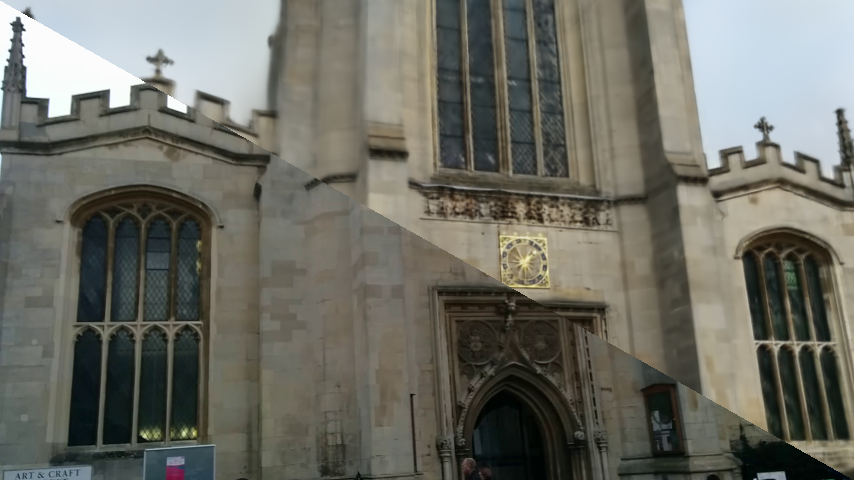}}
\hfill
\subfloat[]{\includegraphics[width=0.24\linewidth]{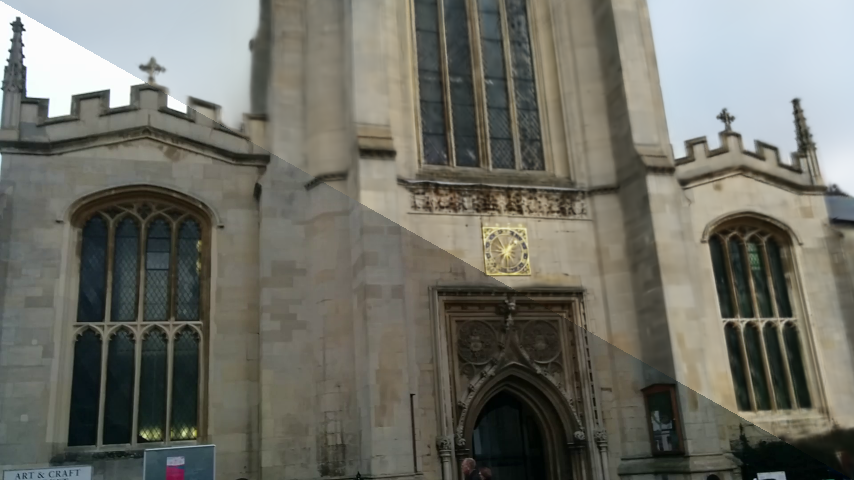}}
\hfill
\subfloat[]{\includegraphics[width=0.24\linewidth]{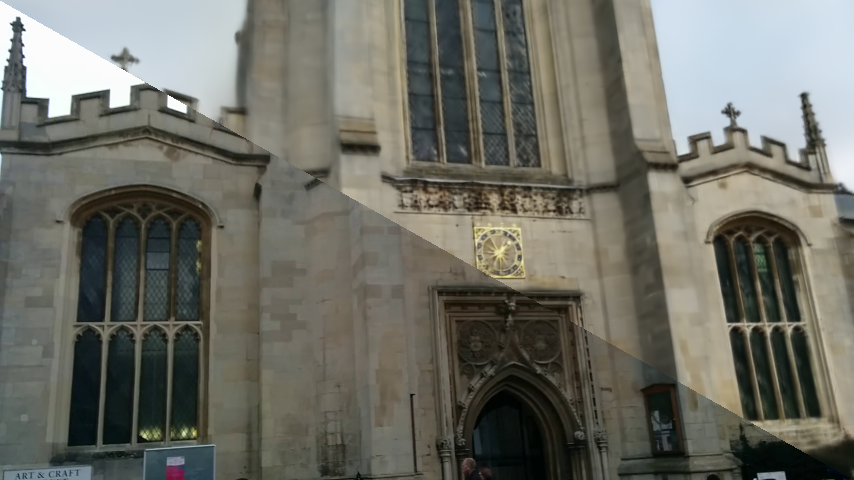}}
\hfill
\subfloat[]{\includegraphics[width=0.24\linewidth]{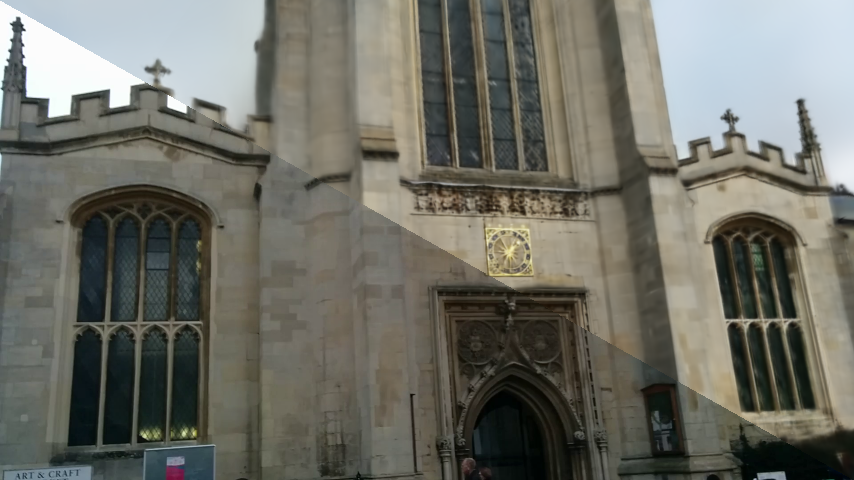}}\\
\captionsetup[subfloat]{labelformat=empty}
\vspace{-0.7cm} 
\subfloat[]{\includegraphics[width=0.24\linewidth]{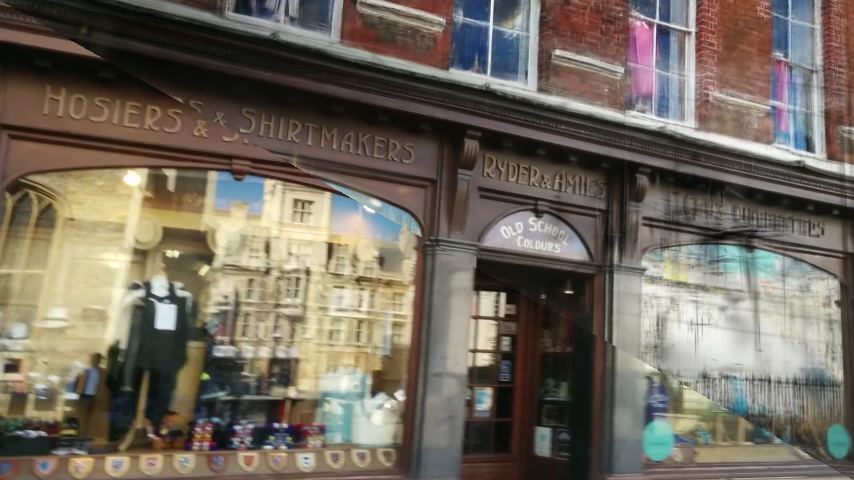}}
\hfill
\subfloat[]{\includegraphics[width=0.24\linewidth]{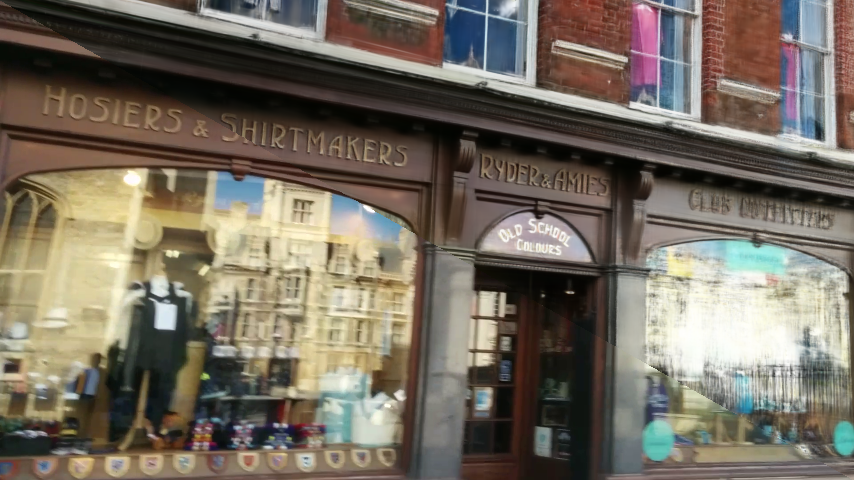}}
\hfill
\subfloat[]{\includegraphics[width=0.24\linewidth]{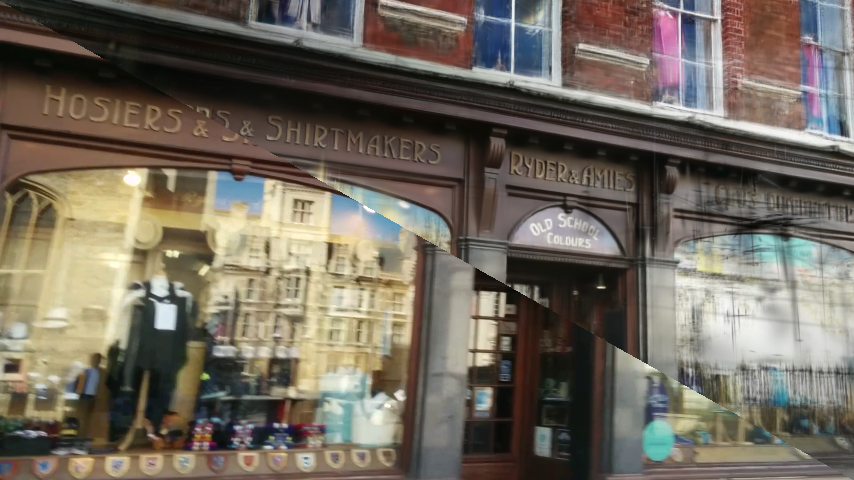}}
\hfill
\subfloat[]{\includegraphics[width=0.24\linewidth]{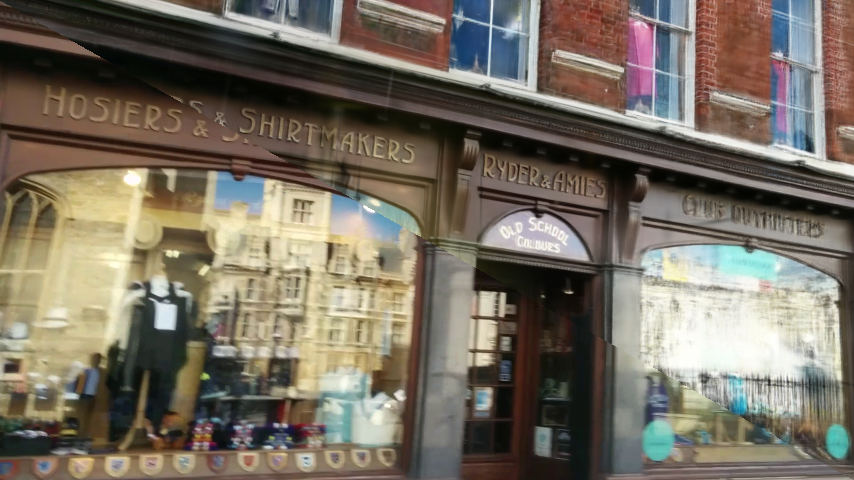}}\\
\captionsetup[subfloat]{labelformat=parens}
\vspace{-0.7cm} 
\setcounter{subfigure}{0} 
\subfloat[DFNet \cite{dfnet}]{\includegraphics[width=0.24\linewidth]{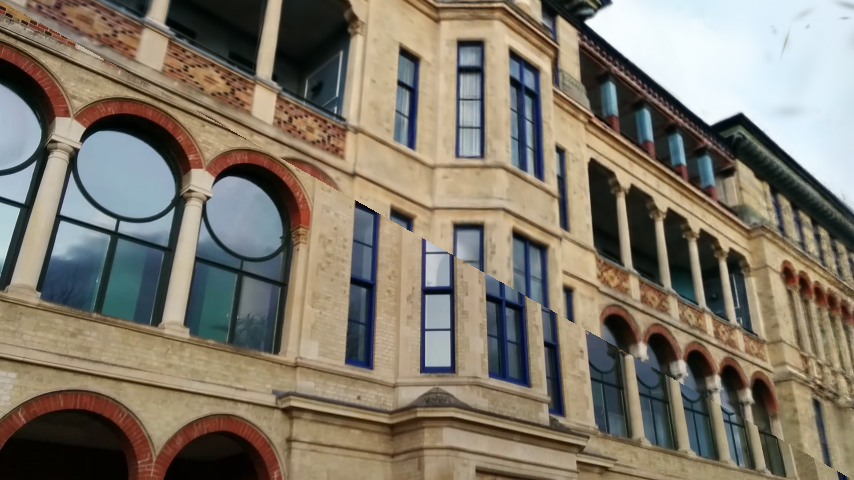}}
\hfill
\subfloat[D.+GS-SMC (ours) \label{fig:d+ours}]{\includegraphics[width=0.24\linewidth]{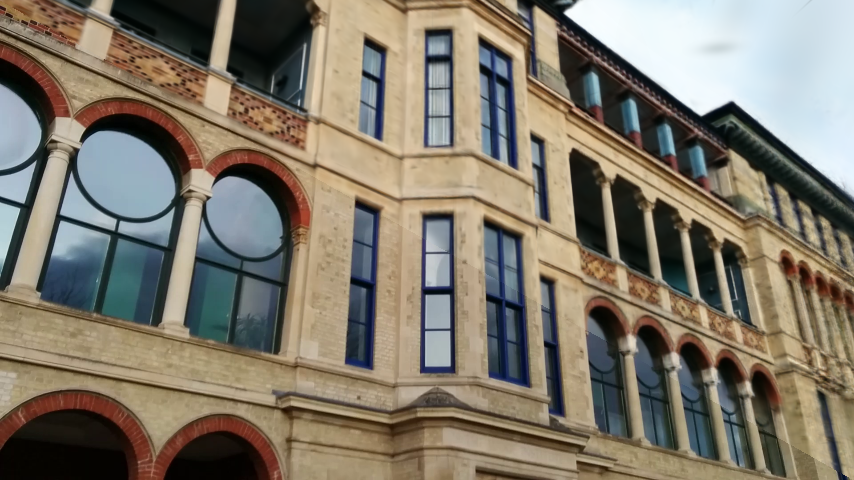}}
\hfill
\subfloat[D.+w/o. multi-view \label{fig:d+w/o multi-view}]{\includegraphics[width=0.24\linewidth]{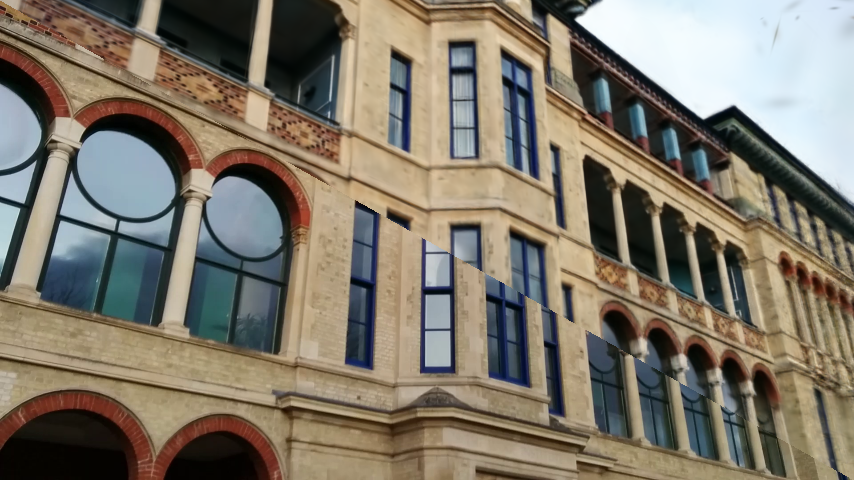}}
\hfill
\subfloat[D.+w/o. EGC \label{fig:d+w/o EGC}]{\includegraphics[width=0.24\linewidth]{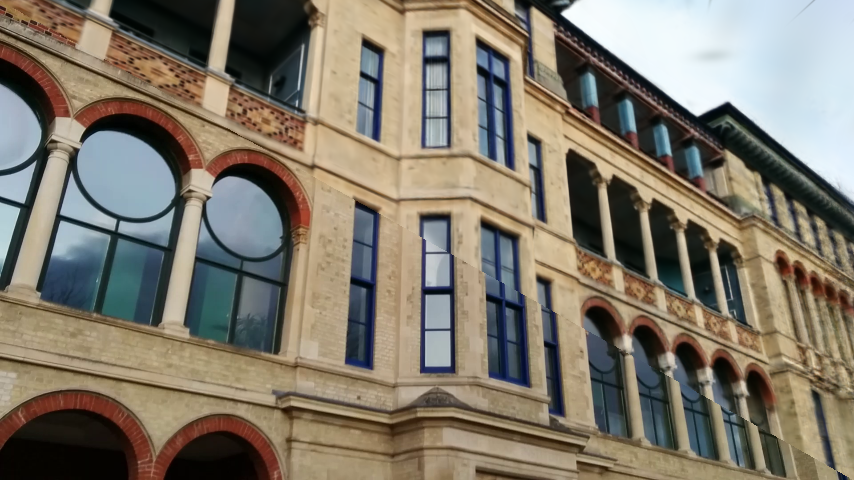}}\\
\captionsetup[subfloat]{labelformat=parens}
\setcounter{subfigure}{0}
\caption{Visualization of GS-SMC improves pose prediction for DFNet on the Cambridge Landmarks dataset. The rows sequentially present the scenes of \textit{Church}, \textit{Shop}, and \textit{Hospital}. Each subfigure is split diagonally, with the bottom left displaying the ground truth query image and the top right showing the rendered estimated/refined pose. }
\label{fig:visualization_cambridge}
\end{figure*}

\begin{figure*}[!t]
    \centering
    \subfloat[Kings]{\includegraphics[width=0.26\linewidth, trim=25 0 0 25, clip]{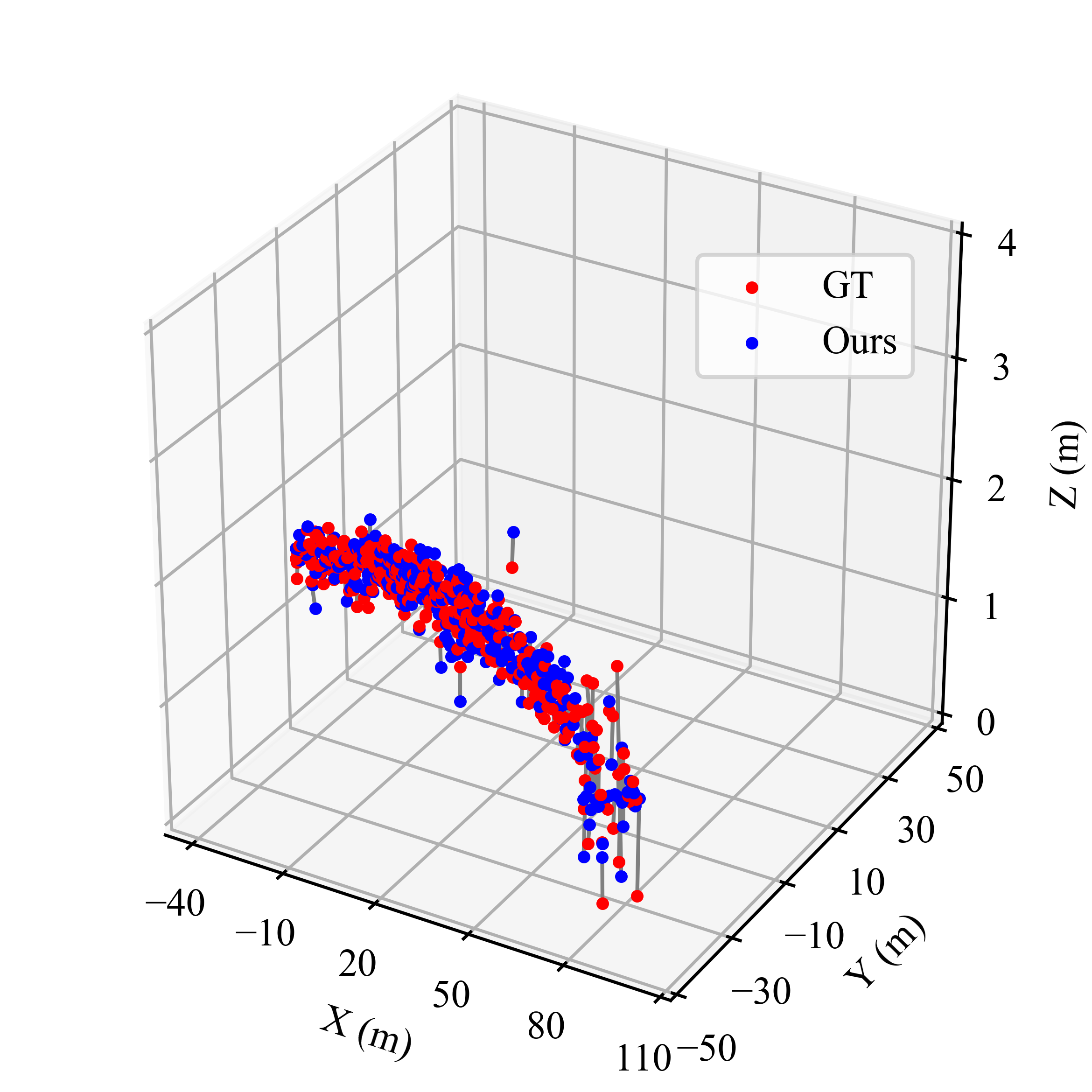}}
    \hspace{0.001\linewidth}
    \subfloat[Church]{\includegraphics[width=0.26\linewidth, trim=25 0 0 25, clip]{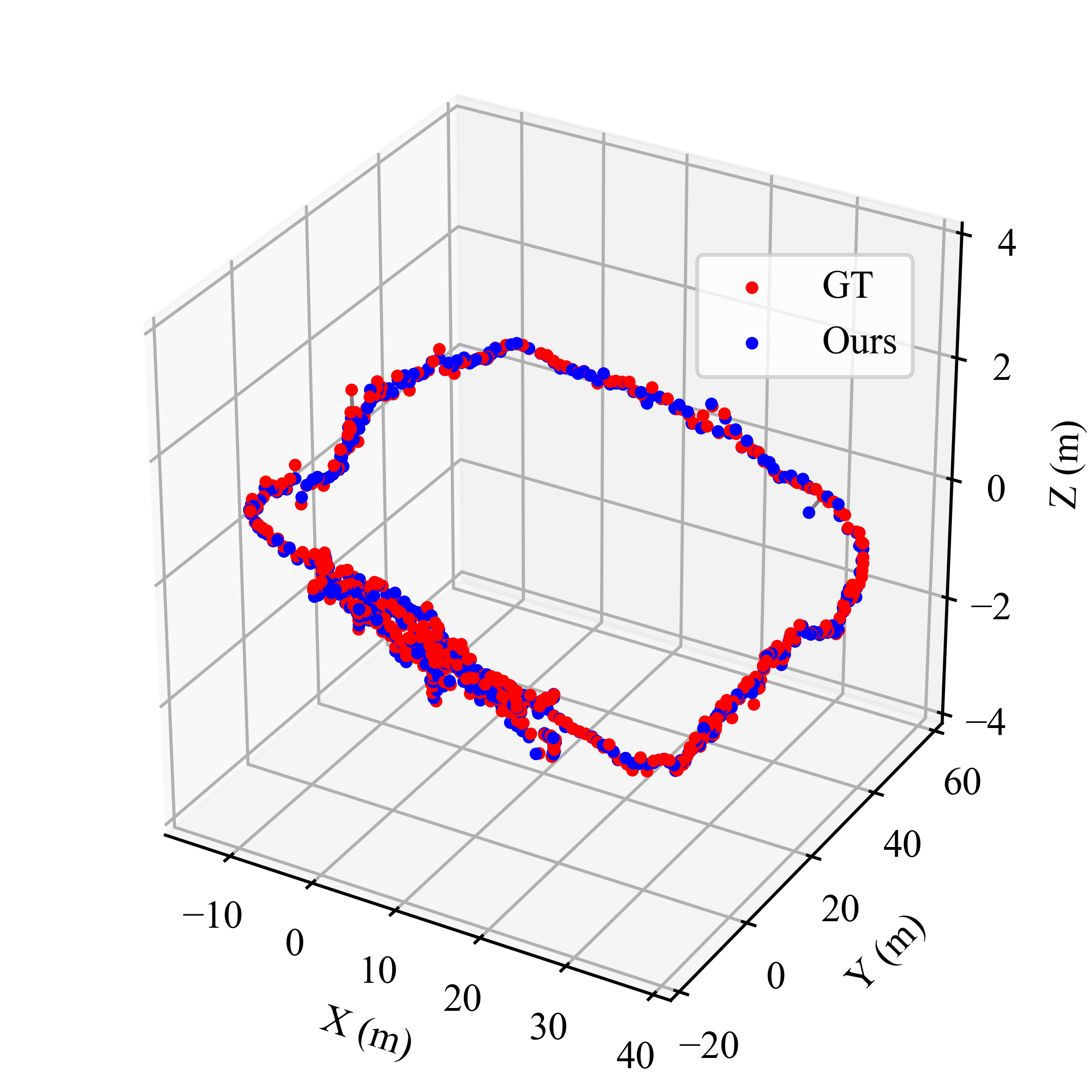}}
    \subfloat[Shop]{\includegraphics[width=0.26\linewidth, trim=25 0 0 25, clip]{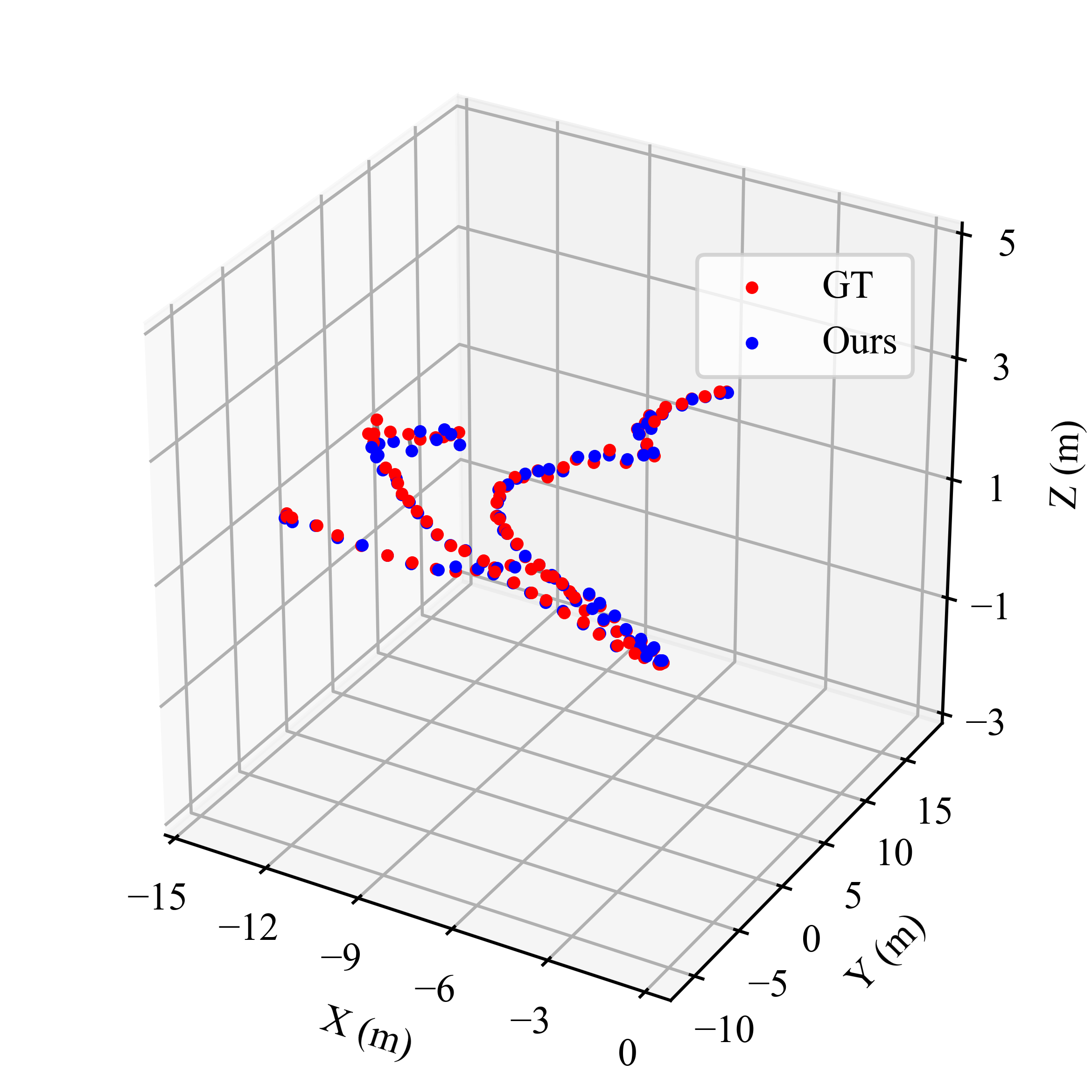}}
    \caption{Predicted camera trajectories for the Cambridge dataset scene, with gray lines indicating translation errors between the estimated and ground truth. \label{fig:visualization_cambridge_trajectory}}
\end{figure*}

Compared to the indoor 7-Scenes dataset, the outdoor Cambridge Landmarks dataset presents more complex and challenging scenes, including lighting variations and moving objects. Similarly, we compare our method with APR methods, including  Ms-Transformer \cite{ms-transformer_2021}, DFNet \cite{dfnet}, LENS \cite{lens}, as well as pose refinement methods, including FQN \cite{FQN}, CrossFire \cite{crossfire}, NeFeS \cite{NeFeS50}, HR-APR \cite{cs-hr-apr}, and MCLoc \cite{mcloc}.

We employ DFNet \cite{dfnet} as the initial pose and iteratively refine it using our proposed GS-SMC to evaluate its effectiveness. The results presented in Table~\ref{tab:Cambridge_output} show that GS-SMC achieves the highest overall performance, surpassing all state-of-the-art APR methods and pose refinement techniques, including MCLoc \cite{mcloc}, NeFeS \cite{NeFeS50}, and HR-APR \cite{cs-hr-apr}, in both translation and rotation accuracy. Specifically, compared to the baseline MCLoc, GS-SMC reduces the average translation and rotation errors by 0.11 m and 0.33$^\circ$ respectively. Moreover, when applied to DFNet as the initial estimates, GS-SMC further improves localization, reducing translation error by 1.03 m and rotation error by 2.61$^\circ$. As illustrated in Fig.\ref{fig:visualization_cambridge}\subref{fig:d+ours}, the rendered image corresponding to the refined pose obtained using our method aligns more closely with the ground truth query image. Moreover, the refined camera trajectories shown in Fig.\ref{fig:visualization_cambridge_trajectory} further demonstrate the robustness and stability of GS-SMC in large-scale outdoor scenes, with the Kings scene exhibiting the largest X-axis span of approximately 150 m.

However, it is worth noting that the complexity of outdoor scenes can sometimes lead to optimization failures. These limitations are discussed further in Sec.~\ref{4.6}.

\begin{table}[!t]
    \captionsetup{justification=centering, labelsep=none} 
    \caption{Ablation experiments \\ on the Cambridge dataset.\label{tab:ablation_without}}
    \centering
    \begin{tabular}{@{\hspace{-1pt}}lcccc}
        \toprule
        \multicolumn{1}{@{\hspace{-1pt}}l}{\multirow{2}{*}{Methods}} & \multicolumn{2}{c}{Hospital} & \multicolumn{2}{c}{Shop} \\
        \cline{2-5} & ${\rm{e}}({\boldsymbol{\hat t}})(m)$ & ${\rm{e}}({\boldsymbol{\hat R}})(^\circ)$ &  ${\rm{e}}({\boldsymbol{\hat t}})(m)$ & ${\rm{e}}({\boldsymbol{\hat R}})(^\circ)$\\ 
        \hline
        DFNet \cite{dfnet}     & 2.0 & 2.98 & 0.67 & 2.21 \\ 
        D.+GS-SMC (ours)     & \textbf{0.25} & \textbf{0.39} & \textbf{0.05} & \textbf{0.24}  \\
        D.+w/o. multi-view      & 1.99 & 1.24 & 0.68 & 0.77  \\ 
        D.+w/o. EGC      & 0.48 & 1.01 & 0.11 & 0.52  \\
        \bottomrule
    \end{tabular}
\end{table}
        
\begin{table}[!t]
    \captionsetup{justification=centering, labelsep=none} 
    \caption{Effects of different noise standard deviations for translation and rotation perturbations on the Cambridge dataset.  \label{tab:ablation_impact}}
    \centering
    \begin{tabular}{@{\hspace{-1pt}}llcccc}
        \toprule
        \multicolumn{1}{{@{\hspace{-1pt}}l}}{\multirow{2}{*}{ }} & \multicolumn{1}{{@{\hspace{-1pt}}l}}{\multirow{2}{*}{ }} & \multicolumn{2}{c}{Hospital} & \multicolumn{2}{c}{Shop} \\
        \cline{3-6} & &  ${\rm{e}}({\boldsymbol{\hat t}})(m)$ & ${\rm{e}}({\boldsymbol{\hat R}})(^\circ)$ &  ${\rm{e}}({\boldsymbol{\hat t}})(m)$ & ${\rm{e}}({\boldsymbol{\hat R}})(^\circ)$\\ 
        \hline
        \multirow{5}{*}{N=6} & $[1,1]$    & 0.29  & 0.48  & 0.06  & 0.24   \\
                            & $[3,4]$    & 0.26  & 0.45   & 0.05  & 0.25   \\
                            & $[4,1]$    & 0.26  & 0.44   & 0.05  & 0.26   \\
                            & $[4,2]$    & \underline{\textit{0.25}} &  \underline{\textit{0.39}} &  \underline{\textit{0.05}} &  \underline{\textit{0.24}} \\
                            & $[4,3]$    & \textbf{0.25} & \textbf{0.38} & \textbf{0.05} & \textbf{0.24}  \\
        \bottomrule
    \end{tabular}
\end{table}

\subsection{Ablation Studies}\label{4.5}

We discuss the validation of the multi-view consistency strategy and epipolar geometry constraints (EGC), as well as the impact of the number of candidate images and initial pose perturbations during iterative refinement. We conduct experiments on the Cambridge Landmarks dataset, using DFNet \cite{dfnet} as the baseline for comparison.

\subsubsection{Verification of Multi-view Consistency Strategy}\label{4.5.1}
To assess the effect of the multi-view strategy, we conducted an ablation study by estimating only the relative pose between the query and reference images. Results, shown in the third row of Table~\ref{tab:ablation_without}, indicate minimal improvement in translation and limited improvement in rotation when the multi-view strategy is omitted. As illustrated in Fig.~\ref{fig:visualization_cambridge}\subref{fig:d+w/o multi-view}, excluding multi-view consistency during refinement substantially degrades performance on outdoor datasets. These findings confirm that incorporating multi-view consistency is essential for accurate pose estimation, particularly in challenging scenarios.

\begin{figure}[!t]
    \centering
    \subfloat[ ]{\includegraphics[width=0.49\linewidth, trim=1 1 10 10, clip]{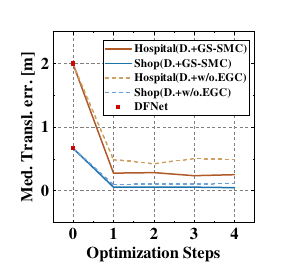}}
    \hspace{0.001\linewidth}
    \subfloat[ ]{\includegraphics[width=0.49\linewidth, trim=1 1 10 10, clip]{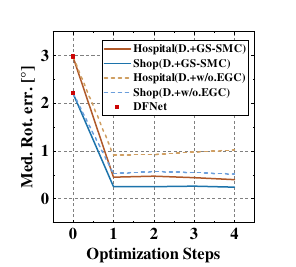}}
    \caption{Optimization Trajectories of DFNet+GS-SMC(ours) and DFNet+without EGC on the Cambridge Dataset, across the scenes of \textit{Hospital} and \textit{Shop}.  \label{fig:visualization_cambridge_iter_abla}}
\end{figure}
            
\begin{figure}[!t]
\centering
\subfloat[ ]{\includegraphics[width=0.49\linewidth, trim=1 1 10 10, clip]{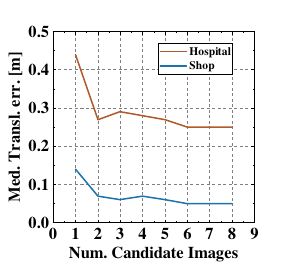}}
\hspace{0.001\linewidth}
\subfloat[ ]{\includegraphics[width=0.49\linewidth, trim=1 1 10 10, clip]{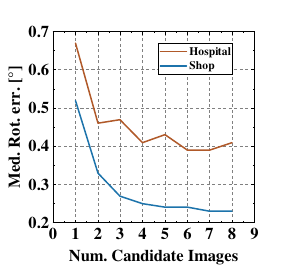}}
\caption{Effects of the number of candidate images on the Cambridge dataset, across the scenes of \textit{Hospital} and \textit{Shop}.  \label{fig:variable_num}}
\end{figure}

\subsubsection{Verification of Epipolar Geometric Constraints} \label{4.5.2}

We conduct experiments to evaluate the effectiveness of Epipolar Geometric Constraints (EGC) by constructing a variant of our GS-SMC framework without it. As shown in Table~\ref{tab:ablation_without}, removing EGC significantly reduces pose estimation accuracy, with localization errors increasing by 0.24 m and 0.62$^\circ$ for the \textit{Hospital} scene, and 0.06 m and 0.28$^\circ$ for the \textit{Shop} scene. Figure~\ref{fig:visualization_cambridge}\subref{fig:d+w/o multi-view} illustrates that the rendered images without EGC align less closely with the query images. Moreover, comparisons in Fig.~\ref{fig:visualization_cambridge_iter_abla} indicate that iterative convergence without EGC is slower and more unstable, whereas our GS-SMC method achieves smoother, more stable, and superior results.

\subsubsection{Variable Number of Candidate Images}\label{4.5.3}

To examine the impact of the number of candidate images, we evaluate the performance across varying numbers of candidates, as shown in Fig.\ref{fig:variable_num}. In this experiment, candidate poses are perturbed with Gaussian noise of 4 m in translation and 2$^\circ$ in rotation. The results show that the translation error remains relatively constant even as more candidates are added, while the median rotation error continues to decrease with additional candidates before eventually converging. Notably, even with only two candidate images, our method still outperforms MCLoc \cite{mcloc}, as shown in Table~\ref{tab:Cambridge_output}.

\subsubsection{Variable Translation and Rotation Perturbations}\label{4.5.4}

Table~\ref{tab:ablation_impact} extends the localization evaluation by varying the standard deviations of Gaussian noise applied to the translation and rotation perturbations of the initial pose, which are then used to render the candidate images. In this experiment, the number of candidate images is fixed at six. Since improving translation accuracy is generally more challenging, we focus on minimizing translation errors. The results indicate that translation perturbations have a greater effect on translation accuracy than rotation perturbations. Furthermore, once the perturbation magnitude exceeds a certain range, increasing it further has only a marginal impact on performance, showing that performance becomes largely insensitive to further increases in the perturbation magnitude.
    
\begin{figure}[!t]
    \centering
    \captionsetup[subfloat]{labelformat=empty}
    \subfloat[]{\includegraphics[width=0.33\linewidth, trim=10 10 10 10, clip]{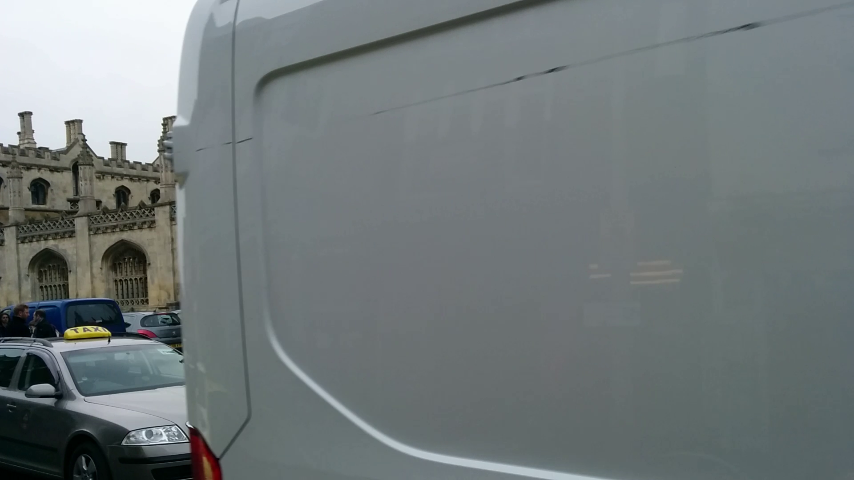}}
    \hfill
    \subfloat[]{\includegraphics[width=0.33\linewidth, trim=10 10 10 10, clip]{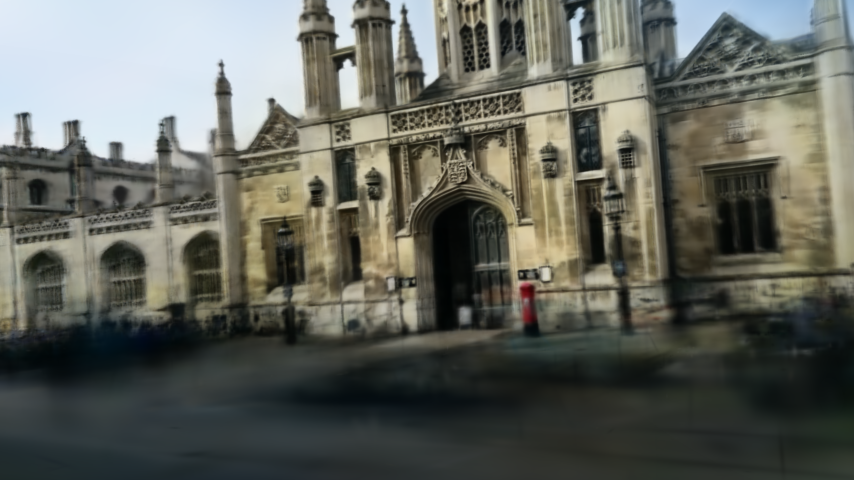}}
    \hfill
    \subfloat[]{\includegraphics[width=0.33\linewidth, trim=10 10 10 10, clip]{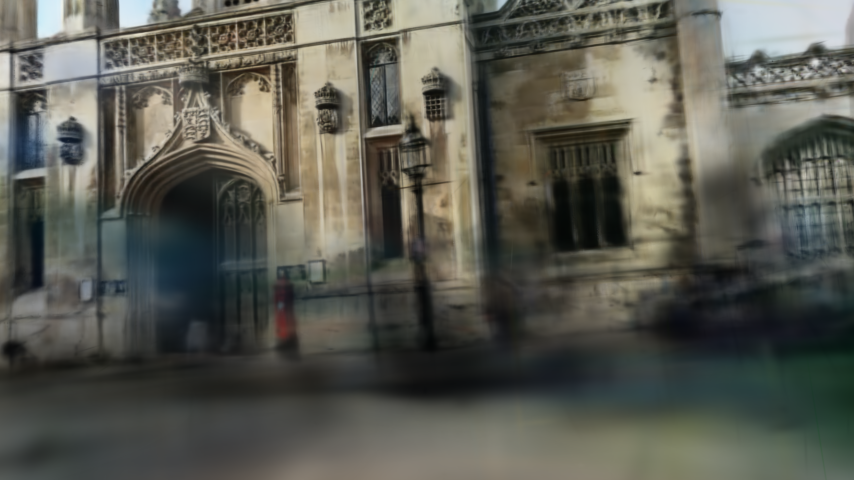}}\\
    \captionsetup[subfloat]{labelformat=parens}
    \setcounter{subfigure}{0} 
    \vspace{-0.7cm} 
    \subfloat[Query]{\includegraphics[width=0.33\linewidth, trim=10 10 10 10, clip]{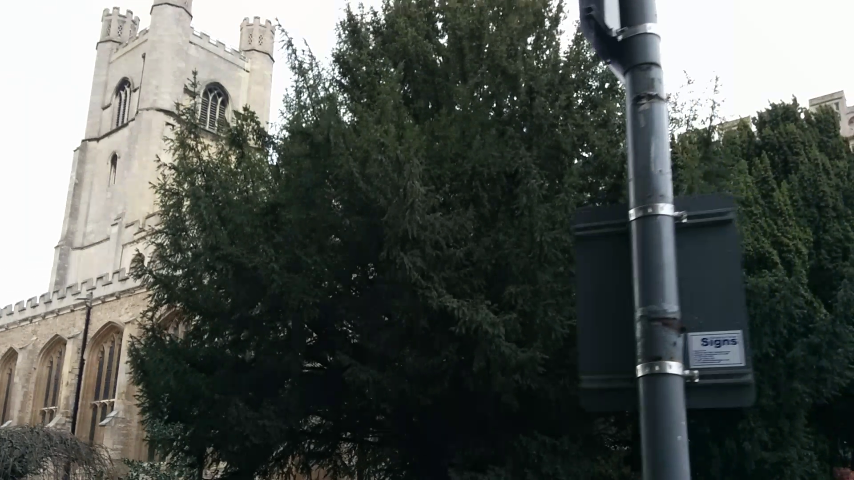}}
    \hfill
    \subfloat[Rendered GT]{\includegraphics[width=0.33\linewidth, trim=10 10 10 10, clip]{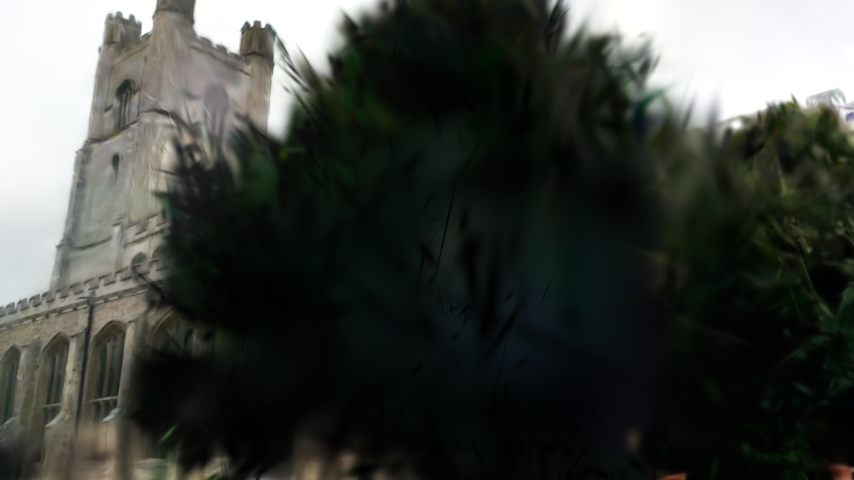}}
    \hfill
    \subfloat[Reference]{\includegraphics[width=0.33\linewidth, trim=10 10 10 10, clip]{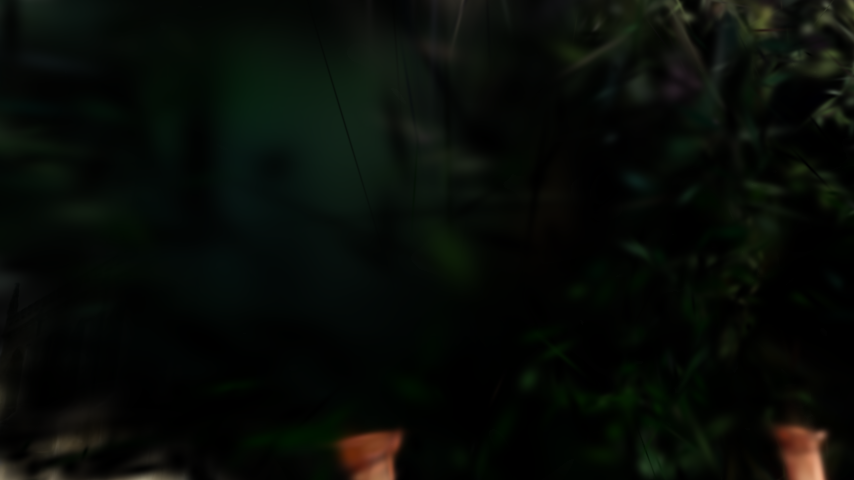}}
    \caption{Visualization of failure cases. \textit{First row}: represents the \textit{Kings}, where the occlusion caused by transient interfering objects. \textit{Second row}: represents the \textit{Church}, where shows the inaccuracy of the initial pose estimation.}
    \label{fig:failure_cases}
\end{figure}

\subsection{Failure Cases and Limitations}\label{4.6}

Currently, our proposed pose refinement approach has some limitations. Fig.~\ref{fig:failure_cases} illustrates observed failure cases in our experiments. One such failure case arises in highly dynamic environments, where rapidly moving objects can occlude visually similar features. This may create discrepancies between rendered and query images, thereby reducing the reliability of feature correspondences. While these challenges mainly stem from scene dynamics, another critical limitation is tied to the quality of the initial pose estimation. Large deviations often prevent convergence and may even mislead the refinement process by producing rendered views that significantly diverge from the real scene structure. Given these inherent limitations in our current framework, we anticipate that our methodology will continue to benefit from advancements in initial pose estimation methods and pre-trained matchers, ultimately resulting in more precise and reliable computations.

\section{Conclusions}

This paper proposes GS-SMC, a novel camera pose refinement framework based on 3D Gaussian Splatting, which eliminates the need for specific descriptors or dedicated neural networks. Specifically, we leverage an existing 3DGS model to generate multiple rendered images, allowing our method to adapt to new scenes without requiring task-specific training or fine-tuning. Furthermore, we propose an iterative refinement approach to estimate the camera pose based on relative poses computed between the query image and the rendered images. Our method supports flexible use of feature extractors and matchers to establish the 2D-2D correspondences required for computing these relative poses. To mitigate the impact of mismatches, we incorporate epipolar geometry constraints with weighted contributions to enhance robustness during the iterative refinement. Extensive experiments demonstrate that the proposed GS-SMC outperforms state-of-the-art pose refinement methods, achieving superior accuracy on both the 7-Scenes dataset and the Cambridge Landmarks dataset.

\bibliographystyle{IEEEtran}
\bibliography{ref}

\vfill

\end{document}